\documentclass[preprint,twocolumn,3p]{elsarticle}
\usepackage{amsmath}
\usepackage{amssymb}
\usepackage{gensymb}%°
\usepackage[]{caption2}
\usepackage{epstopdf}
\usepackage{graphicx}
\usepackage{subfigure}
\usepackage{epsfig}
\usepackage{threeparttable}
\usepackage{graphicx}
\usepackage{lineno,hyperref}
\usepackage{tikz}
\newcommand*{\circled}[1]{\lower.7ex\hbox{\tikz\draw (0pt, 0pt)%
    circle (.5em) node {\makebox[1em][c]{\small #1}};}}
\modulolinenumbers[5]

\journal{Journal of \LaTeX\ Templates}
\begin{document}%

\begin{frontmatter}

\title{A Comparative Study on Movement Feature in Different Directions for Micro-Expression Recognition}
\author[1]{Jinsheng Wei}
\ead{2018010217@njupt.edu.cn}

\author[3]{Guanming Lu\corref{cor}}
\cortext[cor]{Corresponding author}
\ead{lugm@njupt.edu.cn}
\author[2]{Jingjie Yan}

\ead{yanjingjie@njupt.edu.cn}
\address{Nanjing University of Posts and Telecommunications, Nanjing (210003), China}

%\address[mysecondaryaddress]{360 Park Avenue South, New York}pp

\begin{abstract}
Micro-expression can reflect people's real emotions. Recognizing micro-expressions is difficult because they are small motions and have a short duration. As the research is deepening into micro-expression recognition, many effective features and methods have been proposed. To determine which direction of movement feature is easier for distinguishing micro-expressions, this paper selects 18 directions (including three types of horizontal, vertical and oblique movements) and proposes a new low-dimensional feature called the Histogram of Single Direction Gradient (HSDG) to study this topic. In this paper, HSDG in every direction is concatenated with LBP-TOP to obtain the LBP with Single Direction Gradient (LBP-SDG) and analyze which direction of movement feature is more discriminative for micro-expression recognition. As with some existing work, Euler Video Magnification (EVM) is employed as a preprocessing step. The experiments on the CASME \uppercase\expandafter{\romannumeral2} and SMIC-HS databases summarize the effective and optimal directions and demonstrate that HSDG in an optimal direction is discriminative, and the corresponding LBP-SDG achieves state-of-the-art performance using EVM.

\end{abstract}


\end{frontmatter}

\section{Introduction}%PPPpppppppppPp

Facial expression can be divided into macro-expression and micro-expression. Macro-expression is an expression which is observed on the face directly, but when people want to hide their real emotions, we cannot infer their real emotions from macro-expression. When people try to hide their real emotions, micro-expression can represent their real emotions. Although people's real emotions can be inferred from micro-expression, micro-expression cannot be recognized as easily as macro-expression because it is momentary and minor.

Micro-expressions have great research significance and have been studied in the field of psychology for many years\cite{B50,B51,B49,B52,B53}. Micro-expression has been proven to be effective in reflecting people's real emotions, and recognizing micro-expression is valuable for many applications, including lie detection\cite{B1,B54}, medical diagnosis\cite{B53}, public safety\cite{B2} and so on\cite{B55,B56}. Ekman et al. developed the Micro-Expression Training Tool (METT)\cite{B3} to train ordinary people to recognize micro-expressions in seven categories. As micro-expression has minor movements and a momentary duration, recognizing it is artificially difficult. In \cite{B4}, it was shown that undergraduate students who receive the help of METT can only achieve about an accuracy of 40\% in detecting micro-expression. Fortunately, the development of computer vision and pattern recognition has promoted research into automatic micro-expression recognition, and recent research has shown that automatic micro-expression recognition can achieve an excellent recognition rate(from here on, ``micro-expression recognition" below means ``automatic micro-expression recognition").

To study micro-expression recognition more conveniently, many teams have collected micro-expression video databases that are used to verify the effectiveness of their proposed method. Micro-expression databases mainly include spontaneous and non-spontaneous types, and spontaneous databases are more difficult to collect and more realistic than non-spontaneous databases. At present, commonly used spontaneous databases include CASME \uppercase\expandafter{\romannumeral2}\cite{B5}, SMIC\cite{B6}, SAMM\cite{B7} and so on. The recognition rate for these databases has been improved continuously, but micro-expression recognition still faces many challenges, and more effective and robust methods are still needed to promote micro-expression recognition. In this paper, an effective feature is proposed to study movement feature in different directions and improve the recognition rate, and the proposed feature obtains state-of-the-art performance after being concatenated with LBP-TOP.

\section{Related Works}%ppp

This section introduces the related works that have been explored in the field of micro-expression recognition. The recognition of micro-expressions mainly includes three parts: preprocessing, feature extraction and classification. Since the focus of this paper is feature extraction, the following section introduces preprocessing and classification briefly and presents feature extraction in more detail.

In the early stage, some simple explorations and attempts were presented regarding preprocessing and classifiers, but the two parts are relatively fixed. Many databases use video clips where faces have been detected, intercepted and aligned. Eulerian Video Magnification(EVM) \cite{B31} and Time Interpolation Model(TIM)\cite{B30} also are effective preprocessing methods, which have been used widely. The effectiveness of EVM and TIM has been proven in a great deal of work, but they are treated separately. Recently, Peng et al.\cite{B17} combined TIM and EVM to eliminate the side effects caused by the intermediate process and to obtain a state of the art recognition rate. For classifiers, to date, AdaBoost\cite{B57}, Softmax\cite{B27}, KNN\cite{B58} and so on \cite{B15,B23} have been selected, but the most common and effective classifier still is the Support Vector Machine (SVM).

Feature extraction is a key step in micro-expression recognition, and it has always been a research focus. A large number of features have been proposed, such as LBP-TOP, LBP-IP, HIGO, etc. These features can be roughly divided into LBP-based, optical flow-based, gradient-based and deep learning-based methods.
\subsection*{LBP based features}%pppppppppppppP

LBP descriptor was proposed to extract texture feature from two-dimensional images by Ojala\cite{B8}. Zhao et al.\cite{B9} extended LBP from two-dimensional images to three-dimensional videos and obtained LBP-TOP to describe dynamic texture of facial expression video. LBP-TOP is very effective for video analysis, and the LBP-TOP based features have been a research hotspot in micro-expression recognition. A large number of teams have made innovative work. Inspired by the concept of LBP-TOP, Wang et al. proposed more compact LBP-SIP\cite{B10} and LBP-MOP\cite{B11}. LBP-SIP removes redundant points and only calculates the LBP value of six points, and LBP-MOP descriptor does not extract the features from all frames but extracts the features from the average plane. Huang et al.\cite{B12} proposed SpatioTemporal Completed Local Quantization Patterns (STCLQP) and used effective vector quantization and codebook selection to process extracted sign, magnitude and orientation components. After that, they put forward Spatiotemporal Local Binary Pattern with Integral Projection (STLBP-IP\cite{B13}) based on difference image; they calculated the difference images and then extracted LBP features from the integral projection map of the difference images. Huang\cite{B14} and Zong\cite{B15} proposed Discriminative and Hierarchical STLBP-IP respectively to enhance the STLBP-IP feature. Wang et al.\cite{B16} introduced EVM into micro-expression recognition as a preprocessing step and then extracted LBP-TOP to achieve a satisfactory recognition rate. However, their work does not give the recognition rate under the LOSO cross-validation and only tested their method on the CASME \uppercase\expandafter{\romannumeral2} database.

\subsection*{optical flow based features}%pppppppp下・Ppp

Optical Flow (OF) technology was first proposed by Horn et al.\cite{B18} and proven to be effective for micro-expression recognition by several studies. Early, Liong et al.\cite{B20} employed Optical Strain (OS) feature to recognize micro-expressions,  and in the work\cite{B22}, the LBP-TOP was weighted by the temporal mean-pooled OS map in every region; after that, they proposed a Bi-Weighted Oriented Optical Flow (BI-WOOF)\cite{B34} that locally and globally weights HOOF features. To eliminate the influence of noise and illumination changes, Xu et al.\cite{B21} proposed the Facial Dynamics Map (FDM) feature to select the principal direction from the optical flow map. Using the ROI-based OF feature, Liu et al.\cite{B41} presented Main Directional Mean Optical flow (MDMO) to choose the main direction in every ROI. Recently, they\cite{B42} employed sparse representation technology to enhance the feature representation of MDMO and achieve a promising recognition rate. Furthermore, some works combine optical flow and histogram. Zhang et al.\cite{B23} aggregated the Histogram of the Oriented Optical Flow (HOOF) with LBP-TOP features, and inspired by the idea of the fuzzy color histogram, Happy et al.\cite{B40} proposed Fuzzy Histogram of Optical Flow Orientation (FHOFO) that is robust to the variation of expression intensities.
\subsection*{gradient based features}%pppe

The gradient descriptor can extract the movement feature. Histograms of Oriented Gradients (HOG) is an effective gradient feature, and HOG-TOP can be effectively applied to micro-expression recognition after extending HOG to three orthogonal planes. HOG-TOP was first employed by Polikovsky et al.\cite{B24} for micro-expression recognition. In their work, two gradient operators were used to calculate the vertical and horizontal gradients of each pixel in every region of interest (ROI) from three orthogonal planes; then, the gradient direction and gradient magnitude were calculated according to the horizontal and vertical gradients, and the gradient direction weighted by the gradient magnitude was quantified; finally, histogram operation was employed to process these quantized directions. Recently, the histogram of image gradient orientation (HIGO) proposed by Li et al.\cite{B19} does not use the gradient magnitude to weight the gradient direction, and their work employed EVM as a preprocessing method and then achieved an excellent recognition rate.

\subsection*{deep learning based features}%pppppppppp

Recently, deep learning method has been applied in many fields, and Convolutional Neural Network (CNN) is an effective method in the field of image understanding. Because of the limited sample size, CNN model is difficult to be trained for micro-expression recognition. To solve this problem, the pre-trained VGG model and data enhancement technology were employed in the work\cite{B26}, and then the model was fine-tuned to recognize micro-expression using the micro-expression apex frame. Kim et al.\cite{B27} adopted CNN and the Long Short-Term Memory (LSTM) Recurrent Neural Networks to extract spatial features of five states and temporal features respectively. The work\cite{B28} directly use CNN to extract the spatial features of each frame and input these features into LSTM. 3D Convolution Neural Network is a deep learning method for video processing. Li et al.\cite{B29} tried this method in the micro-expression recognition; and in their work, the optical flow maps (horizontal and vertical) and gray-scale frames are gathered and then input into a designed 3DCNN model. Also, the optical flow maps (horizontal and vertical) and the optical strain image are stacked together and input into CNN in work\cite{B39}. The above works promoted the application of deep learning in micro-expression recognition, but their recognition rates are not outstanding compared with traditional methods. By incorporating Accretion Layers (AL) in the network, Verma et al.\cite{B44} proposed a Lateral Accretive Hybrid Network (LEARNet) that refines the salient expression features in accretive manner. Their method achieves a excellent recognition rate, but their experiment don't adopts the mainstream cross-validation method. Furthermore, Khor et al.\cite{B43} proposed a lightweight dual-stream shallow network that has the form of a pair of truncated CNNs with heterogeneous input features, and this method obtains state of the art performance on CASME \uppercase\expandafter{\romannumeral2} database. Recently, using apex frame in micro-expression video, Song et al.\cite{B59} designed a dynamic-temporal stream, static-spatial stream, and local-spatial stream module for the TSCNN that respectively attempt to learn and integrate temporal, entire facial region, and facial local region cues to recognize micro-expressions, and TSCNN achieves the promising recognition. Also, the traditional methods separate feature extraction and classification, while the deep learning methods merge feature extraction and classification and get one model to extract features and classify micro-expressions.
%pp

The optical flow or gradient-based descriptors can extract movement features, but which direction of movement is most conducive for distinguishing micro-expression is still unclear? Extracting the movement feature can be divided into two parts: passive and active. Here, 'passive' means that the extracted movement feature is based on the current video data (for example, in a video clip, if the direction of muscle movement is mainly direction A, the extracted movement feature describes the movement in direction A), while 'active' means extracting the movement feature in a certain direction for all video clips (for example, in all video clips, the movement feature is extracted in the specified direction B.). To study the movement feature in different directions, the 'active' method is more suitable. In existing works, both optical flow and gradient-based descriptors adopted the 'passive' method to extract movement features, so these works cannot be used directly. Besides, according to the definition of optical flow, the optical flow based descriptors cannot be employed to extract movement feature in a specific direction. Thus, we innovated the gradient-based HOG descriptor to meet this demand.

For the studied aim, this paper proposes a new low-dimensional feature called the Histogram of Single Direction Gradient (HSDG) that actively extracts movement feature in a certain direction. The innovation of HSDG is that it simplifies HOG and extracts the movement feature actively. Specifically, on one hand, HSDG only extracts the gradient value in a single direction to ensure that the movement feature in a certain direction are extracted; on the other hand, it does not calculate the gradient direction based on these gradient values, while it directly quantizes these gradient values to reduce the feature dimension. Furthermore, considering that LBP-TOP provides effective appearance texture information, and HSDG cannot provide comprehensive feature information, LBP-TOP, as a basic feature, is concatenated with HSDG to obtain LBP with a Single Direction Gradient(LBP-SDG). The contributions in this paper can be summarized as follows:
%concatenate
\begin{itemize}
  \item [1)]
  In this paper, we study movement feature in different directions and summarize the directions in which movement feature is more useful to distinguish micro-expressions. The experimental results show that the effective direction tends to be the same under different magnification coefficients of EVM.
  \item [2)]
    For the studied aim, a new feature HSDG is proposed. HSDG descriptor can actively extract movement feature in a specific direction. LBP-SDG concatenates HSDG with LBP-TOP and achieves state-of-the-art performance after using EVM.
%Firstly, the gradient in a single direction is calculated, and then these gradient values are quantified into several values, and finally the histogram of the quantified values is calculated.
  \item [3)]HSDG is a low-dimensional feature, and the comparative experiments are used. The results demonstrate that HSDG in the optimal direction is a discriminative feature and can provide effective information. Furthermore, HSDG can be taken as a supplementary feature to improve the performance of basic features.
\end{itemize}

\section{The Proposed Method}
This section introduces our proposed method for the studied aim. The proposed HSDG feature provides effective feature information and achieves a competitive recognition rate after concatenating with LBP-TOP. The proposed method mainly involves preprocessing (EVM, TIM), feature extraction (LBP-TOP, and HSDG). These are described below.
\subsection{Pre-processing}%ppppP

In our work, two effective preprocessing methods (EVM and TIM) were employed, and we introduce the two technologies briefly below.

The difficulties of recognizing micro-expression lie in the fact that micro-expressions are tiny facial movements and have a short occurrence time. Fortunately, EVM can enlarge the muscle movement in the video and has been used widely in micro-expression recognition, so our work also employed this technology. In EVM, the magnification factor $Alpha$ is a key parameter and its value cannot be arbitrarily selected or infinitely large. The optimal $Alpha$ value is different for different databases, so it was fine-selected in the experiment. For detail about EVM, refer to \cite{B31}. Collecting spontaneous micro-expression databases is difficult and the number of collected video frames is uneven. TIM as a normalization method can unify the frames number to solve the problem about the uneven number of frames. TIM was originally proposed for a lipreading system and achieved a very good recognition rate. For detail about TIM, refer to \cite{B30}.

 In our work, the cropped databases that have operated preprocessing such as size normalization, face interception, and face alignment were used directly, and apart from EVM and TIM, no other preprocessing methods are employed.

\subsection{Feature Extraction}%ppp

Feature extraction is the key step of micro-expression recognition and is also the focus of this paper. HSDG is proposed for the studied aim, but the feature information provided by HSDG is incomprehensive. Therefore, HSDG is concatenated with LBP-TOP to obtain LBP-SDG. LBP-TOP and HSDG are introduced in this section.
\begin{figure*}
\begin{center}
\includegraphics[width=1\linewidth]{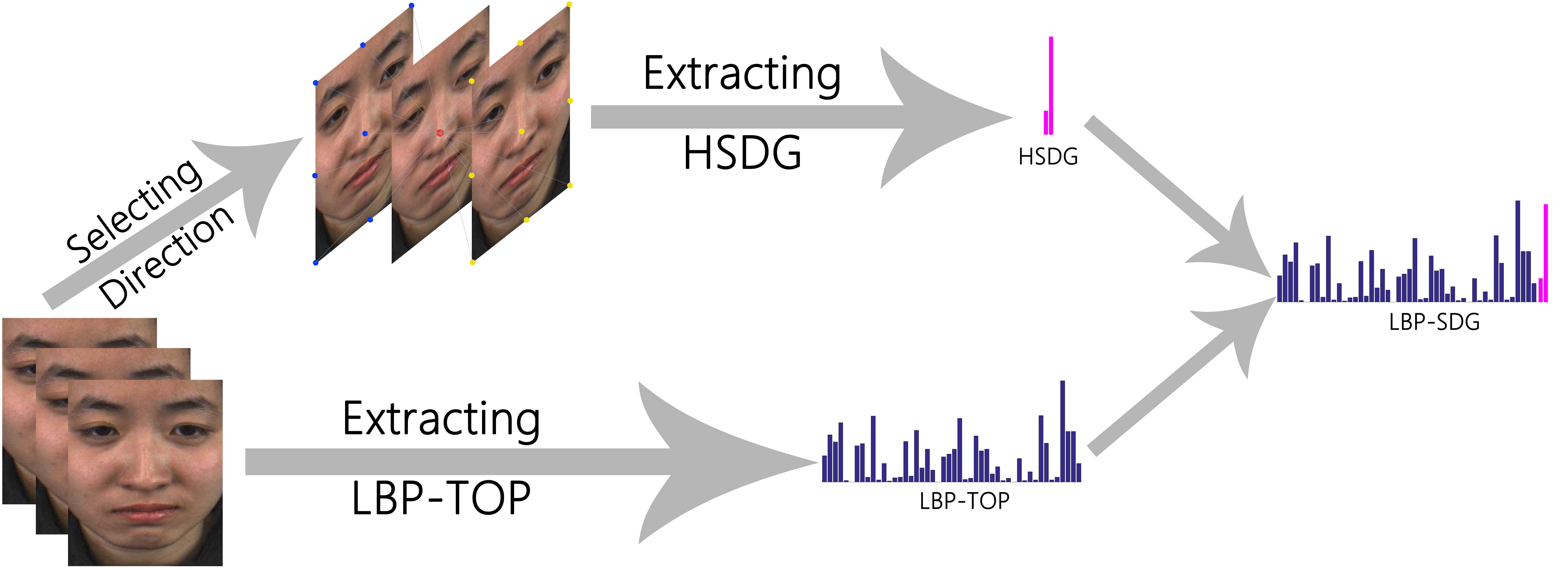}
%\fbox{\rule{0pt}{2in} \rule{.9\linewidth}{0pt}}
\end{center}
   \caption{The framework of extracting LBP-SDG. LBP-TOP and HSDG features are extracted from the micro-expression video clips before these two features are concatenated to obtain LBP-SDG.}
\label{LBPSDG}
\end{figure*}
\subsubsection{LBP-TOP}
A large amount of previous work in different fields shows that LBP-TOP is very effective in video analysis. LBP-TOP can extract texture features from three planes: XY, XT, and YT. The appearance texture features are extracted from the XY plane, while the dynamic texture features are extracted from the other two planes. As LBP is the basis of LBP-TOP, this section first introduces LBP for ease of understanding, and then presents LBP-TOP.

\subsubsection*{LBP}%pp

LBP extracts texture features from two-dimensional images. Suppose $I$ is a two-dimensional image, and a point $P_I=(X_I,Y_I)$ in $I$ is taken as the center point; also, $R_{XI}$ and $R_{YI}$ are the radii on the horizontal and vertical directions, respectively, and $N_I$ points around $P_I$ are selected to calculate the LBP value. First, the coordinates of the $N_I$ points are confirmed, and the coordinates of the i-th point $D_i$ are:

\begin{equation}
\begin{split}
D_i=D(X_{I},Y_{I},R_{XI},R_{YI},N_I,i)=(X_{I}+ R_{XI}\\
*\sin(2{\pi}i/N_I), Y_{I}+R_{YI}*\cos(2{\pi}i/N_I)) \\
i=0,1 ... N_I-1
\end{split}
\label{d}
\end{equation}
These calculated coordinate points form an ordered point set $D_I$. Based on $D_I$, the LBP value of the point $P_I$ is calculated:
\begin{equation}
\begin{split}
LBP(P_{I},D_I)=\sum_{i=0}^{N_I-1}\varepsilon(g(D_i)-g(P_{I}))*2^i
\end{split}
\label{lbp}
\end{equation}

Where $g(x)$ represents the pixel value at point $x$; i represents the index of points in $D_I$; and the $\varepsilon(\ast)$ function is defined as follows:
\begin{equation}
\varepsilon(x)=
\begin{cases}
0& {x<0}\\
1& {x\ge0}
\end{cases}
\end{equation}
Finally, the LBP values of all points on $I$ are calculated, and the histogram of these LBP values is the LBP feature.

\subsubsection*{LBP-TOP}%ppPpp
After presenting the concept of LBP, LBP-TOP is easy to explain. LBP-TOP concatenates these LBP features from the three planes. A three-dimensional video has three coordinate axes (horizontal X, vertical Y, time T), and these three coordinate axes can form three planes (XY, XT, YT). Suppose $V$ is a three-dimensional video, and a point $P_V=(X_V,Y_V,T_V)$ in $V$ is taken as the center point; also, $R_{XV}$, $R_{YV}$, and $R_{TV}$ are the radii in the horizontal, vertical and time directions, respectively, and $N_{XYV}$, $N_{XTV}$ and $N_{YTV}$ points around $P_V$ are selected to calculate the LBP values from the plane XY, XT, and YT, respectively. First, the coordinates of the i-th point ($D_{XYi}$, $D_{XTi}$, and $D_{YTi}$) on the three planes are as follows:

\begin{subequations}
\begin{equation}
\begin{split}
D_{XYi}=D(X_V,Y_V,+R_{XV},-R_{YV},N_{XYV},i)
\end{split}
\end{equation}
\begin{equation}
\begin{split}
D_{XTi}=D(X_V,T_V,+R_{XV},+R_{TV},N_{XTV},i)
\end{split}
\end{equation}
\begin{equation}
\begin{split}
D_{YTi}=D(Y_V,T_V,-R_{YV},+R_{TV},N_{YTV},i)
\end{split}
\end{equation}
\end{subequations}%

where function $D(*,*,*,*,*)$ mentioned in Equation~\ref{d}; $XYi$ = 0,1,...,$N_{XYV}-1$; $XTi$ = 0,1,...,$N_{XTV}-1$; and $YTi$ = 0,1,...,$N_{YTV}-1$.

All $D_{XYi}$, $D_{XTi}$, and $D_{YTi}$ over all i form three ordered point sets, $D_{XY}$, $D_{XT}$, and $D_{YT}$ , respectively. Next, the LBP values ($LBP_{XY}$, $LBP_{XT}$, and $LBP_{YT}$) of point $P_V$ on the three planes are determined by the following equation:
\begin{subequations}
\begin{equation}
\begin{split}
LBP_{XY}(P_V)=LBP(P_V,D_{XY})
\end{split}
\end{equation}
\begin{equation}
\begin{split}
LBP_{XT}(P_V)=LBP(P_V,D_{XT})
\end{split}
\end{equation}
\begin{equation}
\begin{split}
LBP_{YT}(P_V)=LBP(P_V,D_{YT})
\end{split}
\end{equation}
\end{subequations}
Where function $LBP(*,*)$ mentioned in Equation~\ref{lbp}.

Finally, these LBP features from three planes are calculated and concatenated to get LBP-TOP.

\subsubsection{HSDG}%pp

The studied aim in this paper is to determine which directions of movement feature are most effective, so we need to actively extract movement features in a single direction. Considering the gradient feature can extract the movement feature as described by \cite{B25}, the proposed HSDG feature employs the gradient method and extracts gradient features in a single direction. Besides, the proposed feature is inspired by HOG features to calculate gradient values in a single direction and then the histogram of these gradient values. Furthermore, before calculating the histogram, these gradient values are quantified into several values to reduce the feature dimension.

The differences between HOG/HIGO and HSDG are as follows: first, HOG/HIGO calculates the gradient direction according to the calculated horizontal and vertical gradient values, while HSDG calculates the gradient value in only a certain direction. Second, HOG/HIGO quantifies the gradient direction into several directions, while HSDG quantizes the calculated gradient value into several values. Third, overall, HOG/HIGO calculates the gradient direction passively, while HSDG calculates the gradient values in a single direction actively. The specific details of HOG/HIGO can be found in \cite{B32} and \cite{B19}, and the method of calculating HSDG is as follows.

\subsubsection*{Directions}%PPPpppppppppp

\begin{figure}[t]
\begin{center}
\includegraphics[width=0.94\linewidth]{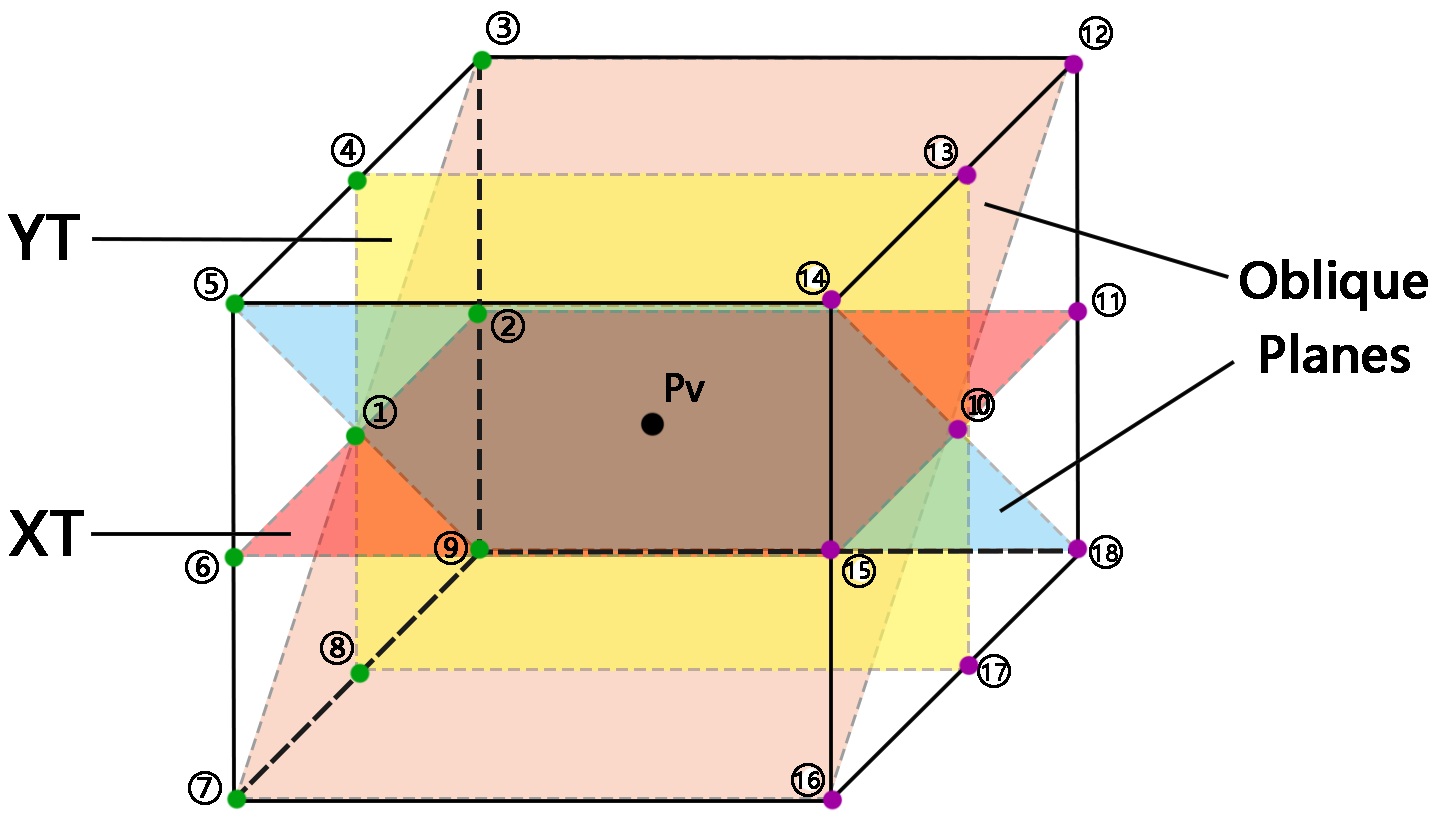}
\end{center}
\caption{The selected 18 directions. $P_V$ represents the center point. The directions \protect\circled{1}, \protect\circled{2}, \protect\circled{6}, \protect\circled{10}, \protect\circled{11} and \protect\circled{15} lie the horizontal plane (XT), the directions \protect\circled{4}, \protect\circled{8}, \protect\circled{13} and \protect\circled{17} lie the vertical plane (YT), and the other directions lie the oblique planes.}
\label{directions}
\end{figure}

Before introducing HSDG, the 18 directions were selected for testing. In $V$, we select $P_V=(X_V,Y_V,T_V)$  as the center point, and $R_{X}$, $R_{Y}$ and $R_{T}$ are the radii in the horizontal, vertical and time directions, respectively. The point \circled{i} is selected in $V$, and i is from 1 to 18. As the extracted feature is used to describe movement information, $R_{T}$ must not be equal to 0; that is, the points should be selected from the previous or subsequent frames. As shown in Figure~\ref{directions}, six points, four points, and eight points are selected on the horizontal (XT), vertical (YT), and oblique planes, respectively; for example, point \circled{11} belongs to the horizontal plane. Then, based on these points, $\vec{P_V\circled{i}}$ is selected as the tested direction. Since the directions are numbered as the same order of point, and \circled{i} is used directly to represent the direction in the subsequent text. The selected 18 points form a point set D:
\begin{equation}
\begin{split}
D=\{
(X_{V}+0\ \ \  ,Y_{V}+0\ \ \ ,T_{V}-R_{T}),\\
(X_{V}-R_{X},Y_{V}+0\ \ \ ,T_{V}-R_{T}),\\
(X_{V}-R_{X},Y_{V}+R_{Y},T_{V}-R_{T}),\\
(X_{V}+0\ \ \ ,Y_{V}+R_{Y},T_{V}-R_{T}),\\
(X_{V}+R_{X},Y_{V}+R_{Y},T_{V}-R_{T}),\\
(X_{V}+R_{X},Y_{V}+0\ \ \ ,T_{V}-R_{T}),\\
(X_{V}+R_{X},Y_{V}-R_{Y},T_{V}-R_{T}),\\
(X_{V}+0\ \ \ ,Y_{V}-R_{Y},T_{V}-R_{T}),\\
(X_{V}-R_{X},Y_{V}-R_{Y},T_{V}-R_{T}),\\
(X_{V}+0\ \ \ ,Y_{V}+0\ \ \ ,T_{V}+R_{T}),\\
(X_{V}-R_{X},Y_{V}+0\ \ \ ,T_{V}+R_{T}),\\
(X_{V}-R_{X},Y_{V}+R_{Y},T_{V}+R_{T}),\\
(X_{V}+0\ \ \ ,Y_{V}+R_{Y},T_{V}+R_{T}),\\
(X_{V}+R_{X},Y_{V}+R_{Y},T_{V}+R_{T}),\\
(X_{V}+R_{X},Y_{V}+0 \ \ \ ,T_{V}+R_{T}),\\
(X_{V}+R_{X},Y_{V}-R_{Y},T_{V}+R_{T}),\\
(X_{V}+0\ \ \ ,Y_{V}-R_{Y},T_{V}+R_{T}),\\
(X_{V}-R_{X},Y_{V}-R_{Y},T_{V}+R_{T})\}
\end{split}
\end{equation}

Theoretically, the movement features between directions \circled{1}-\circled{9} and directions \circled{10}-\circled{18} should be consistent. For example, the movement features in direction \circled{7} and direction \circled{12} should be similar, since the two directions are on the same line. However, since HSDG and LBP-TOP adopt the same method of block division for consistency and the efficiency of the extracted feature, the points in the front and rear $R_{TV}$ frames of a video cannot be taken as the center point. Thus, the points in the rear $R_{TV}$ frames cannot be used when extracting HSDG in directions \circled{1}-\circled{9}, while the points in the former $R_{TV}$ frame cannot be used for extracting HSDG in directions \circled{10}-\circled{18}. Therefore, the movement features in directions \circled{1}-\circled{9} and directions \circled{10}-\circled{18} are different and are tested separately.

\subsubsection*{Extraction Detail}%0p

We pick a direction, and the HSDG features in the direction can then be extracted. First, we select a point P from D; then, the gradient in the direction $\vec{P_V P}$ is:
\begin{equation}
G(\vec{P_VP})=g(P)-g(P_V)
\label{g}
\end{equation}

As the pixel value ranges from 0 to 255, the gradient value ranges from -255 to 255. Considering that the histogram operation on all gradient values(511 types) will generate a huge feature dimension, all gradient values are quantized to N values. All gradient values are quantified by the following equation:
\begin{equation}
q(x)=
\begin{cases}
0& {-255\le x < f(-255+\frac{511*1}{N})}\\
1& {f(-255+\frac{511*1}{N}) \le x< f(-255+\frac{511*2}{N})}\\
... \ & \ ...\\
N-1& {f(-255+\frac{511*(N-1)}{N})\le x \le 255}
\end{cases}
\label{q}
\end{equation}

where f(x) is Rounddown function, and the input x of q(x) is the gradient value.

After the quantized gradient values at all points are calculated, the histogram of these values is the HSDG features.

When a facial muscle movement appears in an area, it will cause corresponding facial movement variations in the area and its surrounding areas. These variations are reflected in the video data as the variations of the pixel values. The proposed HSDG descriptor can extract movement features in a certain direction, and HSDG feature can express the variations of the pixel values in a certain direction. 18 representative directions were selected and the planes where these directions lie include horizontal, vertical, and oblique planes. Among these tested directions, the effective direction means that the movement features (the variations of the pixel values) in this direction are discriminative for distinguishing micro-expressions. In other words, facial movement variations in this direction are good for recognizing micro-expressions. It is noted that directions \circled{1} and \circled{10} are the directions along the time axis, and HSDG features in the two directions express the variations of the pixel values in the same pixel at different times.

\subsubsection{LBP-SDG}%pp

HSDG descriptor can meet the studied aim in this paper, but the movement feature extracted by it is only in a single direction and provides limited feature information. LBP-TOP descriptor can extract appearance texture and dynamic texture features and provides relatively sufficient texture feature information. Thus, HSDG is concatenated with LBP-TOP to obtain LBP-SDG for studying the movement features in different directions. Taking the recognition rate of LBP-TOP as a reference, when the HSDG feature in a certain direction is added, if the recognition rate of LBP-TOP is improved, the movement feature in this direction can provide effective information. The entire feature extraction process is shown in Figure~\ref{LBPSDG}.

\section{Experiments}%PPp
A large number of experimental results and the analysis are shown in this section. Our experiments tested LBP-SDG in 18 directions to summarize the directions in which movement feature is effective for recognizing micro-expressions. If the recognition rate of LBP-SDG in a certain direction is higher than that of LBP-TOP, the movement feature provided by the corresponding HSDG is effective; that is, the movement feature in this direction is effective, and vice versa. LBP-SDG in an optimal direction was compared with other features to prove that LBP-SDG is effective, and HSDG is a discriminative feature to recognize micro-expression. Furthermore, comparing with other methods, EVM+LBP-SDG has state-of-the-art performance.

The relative recognition rate is mentioned in this paper for the most convenient description, where the relative recognition rate is the recognition rate of the LBP-SDG minus that of LBP-TOP. Thus, the effective directions are the directions in which the relative recognition rate is greater than 0. Note that the radius parameters ($R_{XV}$, $R_{YV}$, $R_{TV}$, $R_{X}$, $R_{Y}$, and $R_{T}$) determine the time step and gradient value, so the study is based on these parameters.

\subsection{Setting}%PppP

\begin{figure*}[htb]%
\begin{center}
\subfigure [The recognition rate under not using EVM.]{
\label{SnoX}
\includegraphics[height=5.5cm,width=7.9cm]{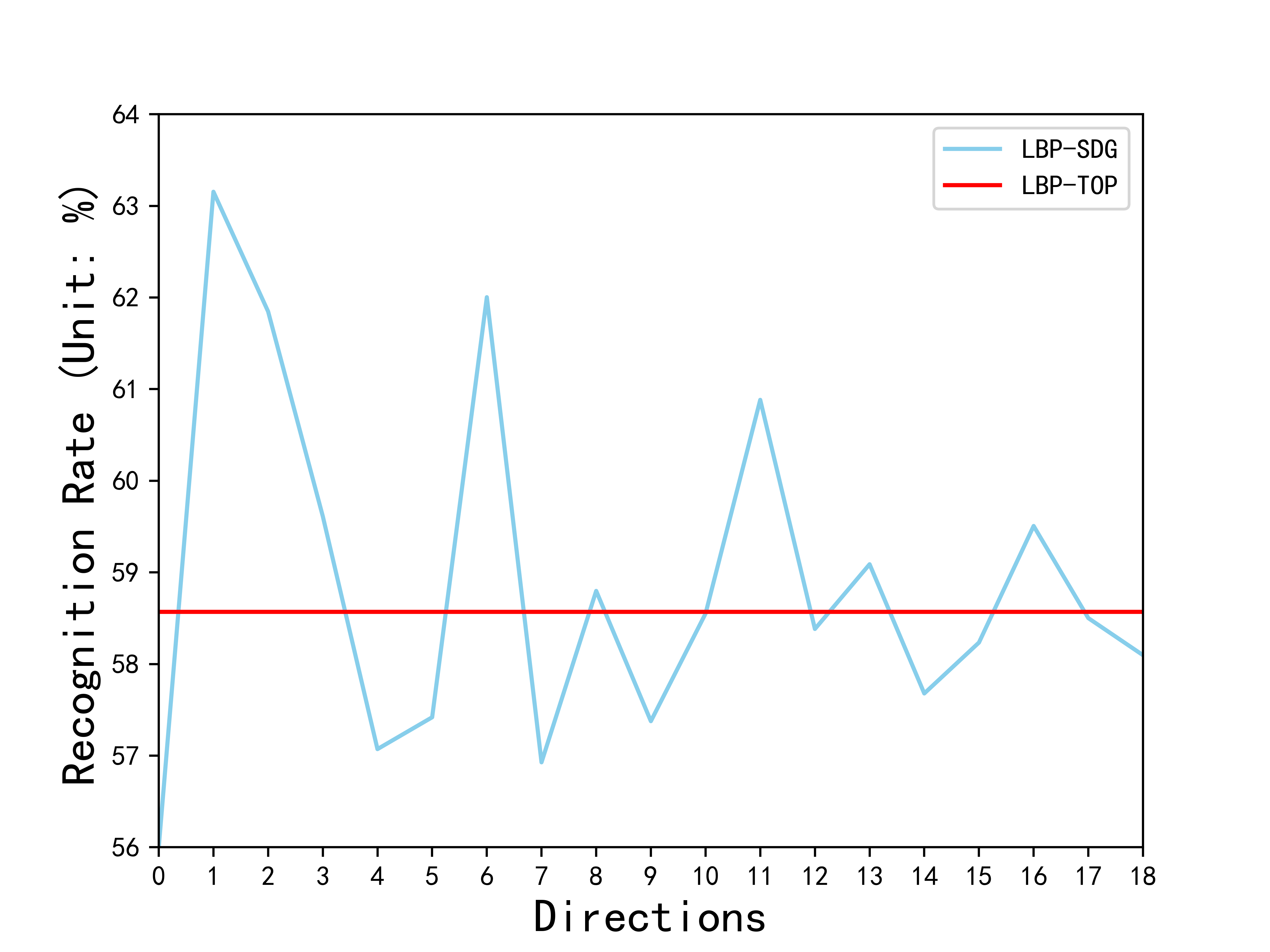}
}
\subfigure [The relative recognition rate under not using EVM.]{
\label{SnoP}
\includegraphics[height=5.5cm,width=7.9cm]{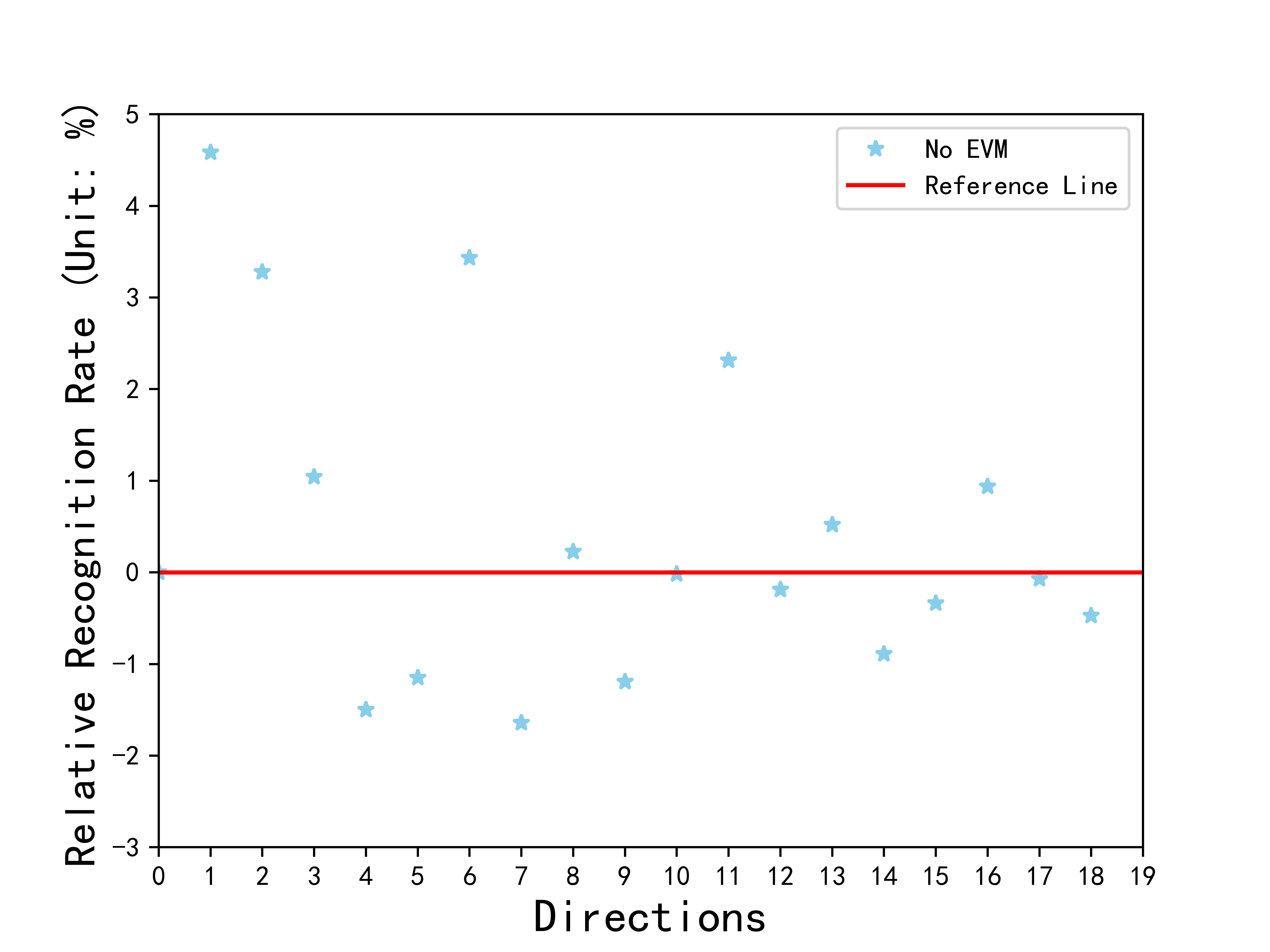}
}

\subfigure [The recognition rate under using EVM.]{
\label{SX}
\includegraphics[height=5.5cm,width=7.9cm]{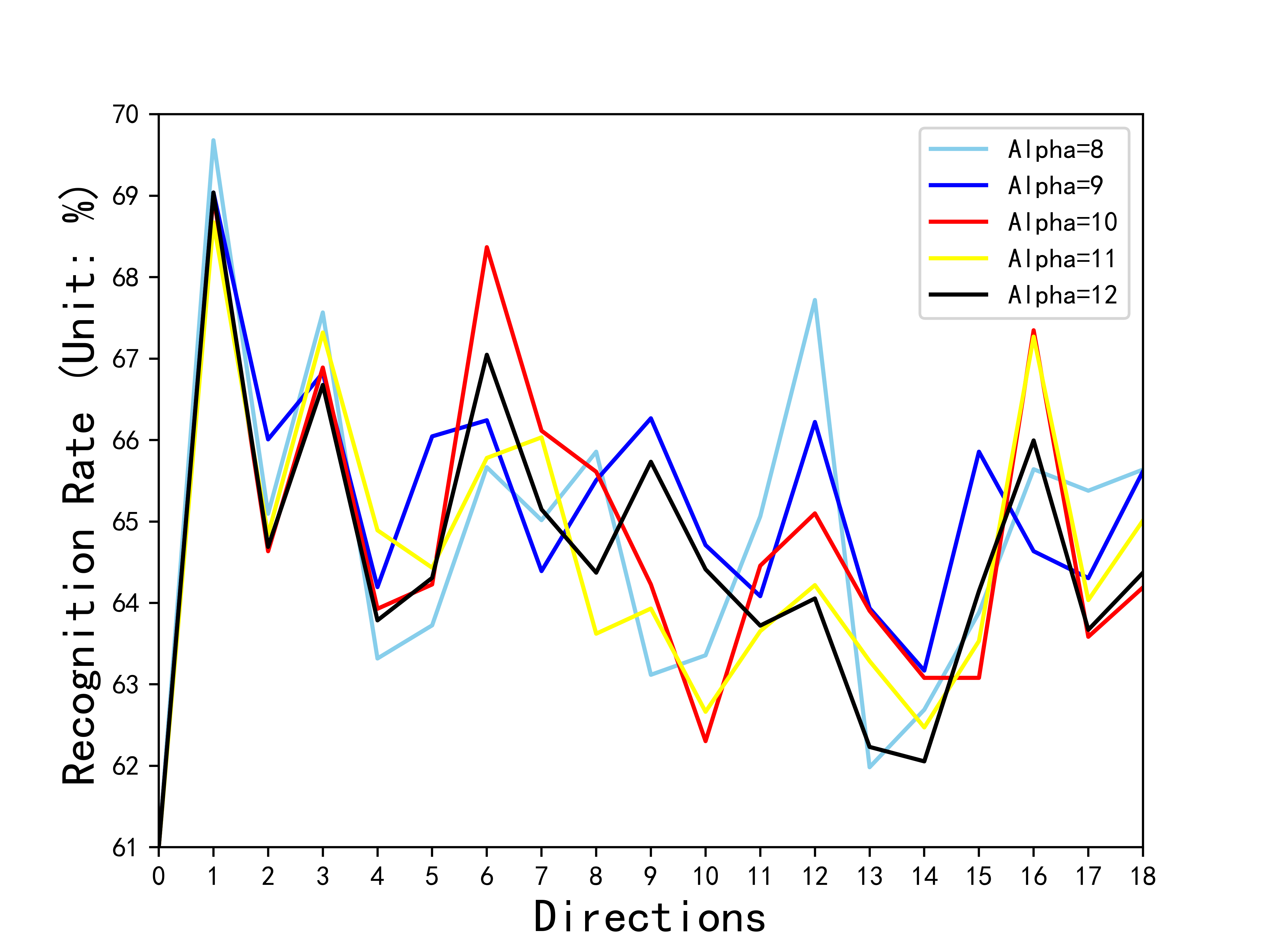}
}
\subfigure [The relative recognition rate under using EVM.]{
\label{SP}
\includegraphics[height=5.5cm,width=7.9cm]{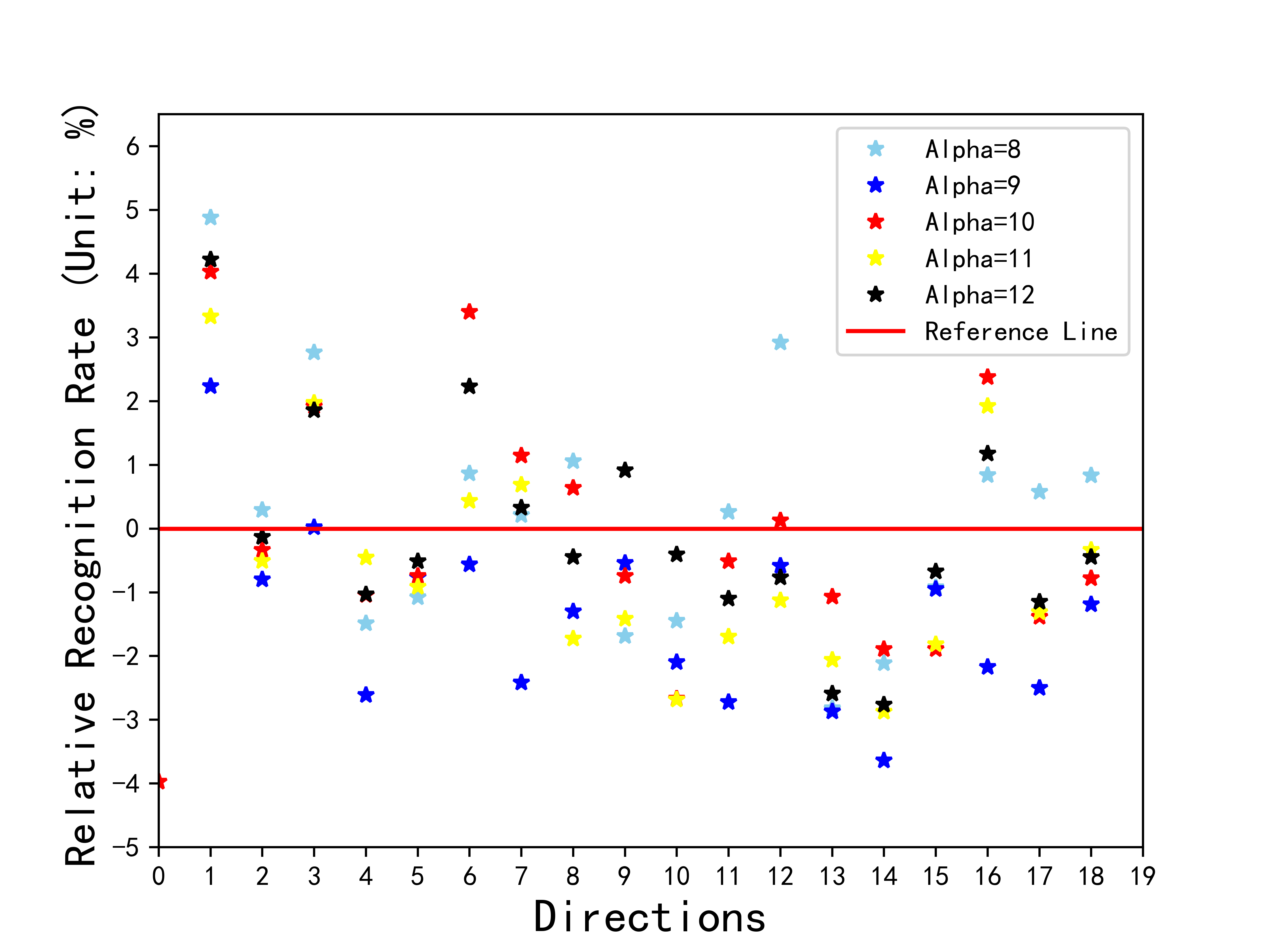}
}
\caption{The SMIC-HS database: the line chart of the recognition rate and the dot charts of relative recognition rate in 18 directions. }
\label{SF}
\end{center}
\end{figure*}

All experiments reported the recognition rate (RR) under leave-one-subject-out (LOSO) cross-validation. Experiments include the following three parts: 1) LBP-SDG in 18 directions was tested for the studied aim; 2) LBP-SDG was compared with other features under the same conditions; 3) LBP-SDG in the optimal direction was compared with other methods (including the state of the art method) by showing their recognition rate directly. All experiments were performed on SMIC and CASME \uppercase\expandafter{\romannumeral2} databases.

The SMIC database is the first spontaneous micro-expression database published by the University of Oulu, and the participants come from multiple countries. The SMIC database includes three subsets collected by three types of cameras: highspeed (HS), normal visual (VIS), and near-infrared (NIS). The SMIC-HS database is a subset of SMIC with the largest sample number and frame rate and was adopted in our experiment. The SMIC-HS database contains 164 samples and is divided into three categories: ``Positive", ``Negative" and ``Surprise". Support Vector Machine with the linear kernel was employed as the classifier. The block strategy was adopted, and each video was divided into 8 * 8 * 2 blocks\cite{B38}. The frames number was unified to 10 using TIM and each frame was resized to 168 * 136 (grayscale image); thus, each video was normalized to 168 * 136 * 10. The quantized number N was set to 2, which is equivalent to binarization according to Equation~\ref{q}. Also, $R_V = R_{XV} = R_{YV} = R_{TV} = 1, 2, 3, 4$, $R = R_{X} = R_{Y} = R_{T} = 1, 2, 3, 4$, and $Alpha = 8, 9, 10, 11, 12$.

The CASME \uppercase\expandafter{\romannumeral2} database is the largest spontaneous micro-expression database published by the Institute of Psychology, Chinese Academy of Sciences, and the participants come from China. The CASME \uppercase\expandafter{\romannumeral2} database contains 255 samples (26 subjects) and is divided into seven categories. The two types of samples ( ``Sadness" and ``Fear" expressions) are relatively small compared to other types, so they are deleted. Eventually, five types of expression samples (``Happiness", ``Disgust", ``Repression", ``Surprise", and ``Others") are selected, and the selected sample size is 246. Support Vector Machine with the Chi-Square kernel was employed as the classifier. Each video was divided into 5 * 5 * 1 blocks. The number of frames adopts the raw number without using TIM, and each frame was un-resized to unify size. The quantized number N was set to 2. $R_{XV} = R_{YV} = 1$, $R_V = R_{TV} = 2, 3, ..., 8$, $R_{X} = R_{Y} = 1$, $R = R_{T} = 2, 3, ... , 8$, and $Alpha = 17, 20, 23, 26, 29$. Also, the uniform LBP\cite{B37} was employed to calculate LBP-TOP.

As for the determination of alpha values, first, we tested the alpha values (from 1 to 30) for LBP-TOP, and found the optimal alpha value for LBP-TOP; second, we selected the representative alpha values around the optimal alpha value (including the optimal alpha value), and under these selected alpha values, the recognition rate of LBP-TOP first increases and then decreases; finally, different features were verified under these selected alpha values.

\subsection{The Study of Movement Feature in Different Directions}%PppPPp

\begin{figure*}[htb]
\begin{center}
\subfigure [The recognition rate under not using EVM.]{
\label{CnoX}
\includegraphics[height=5.5cm,width=7.9cm]{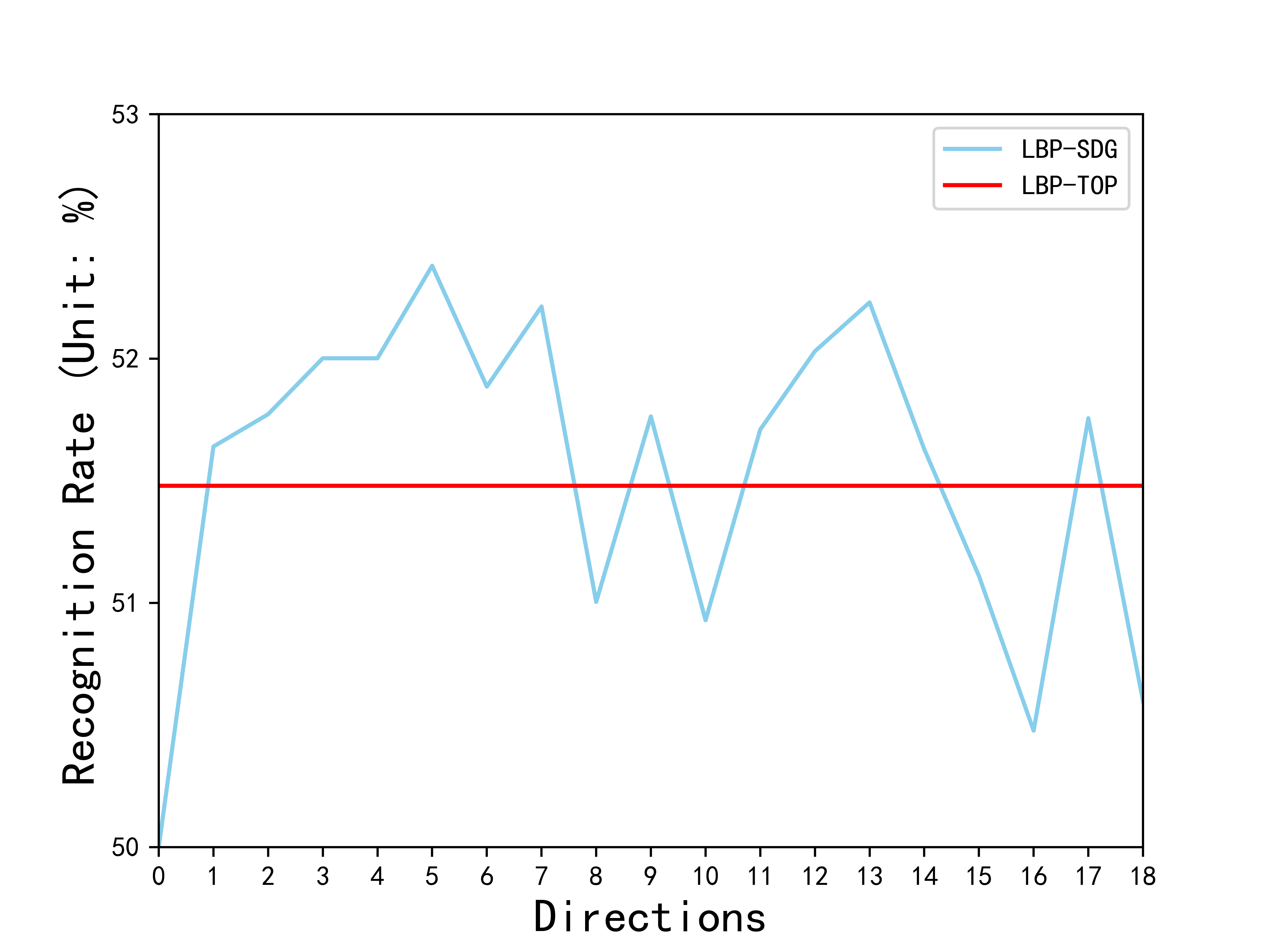}
}
\subfigure [The relative recognition rate under not using EVM.]{
\label{CnoP}
\includegraphics[height=5.5cm,width=7.9cm]{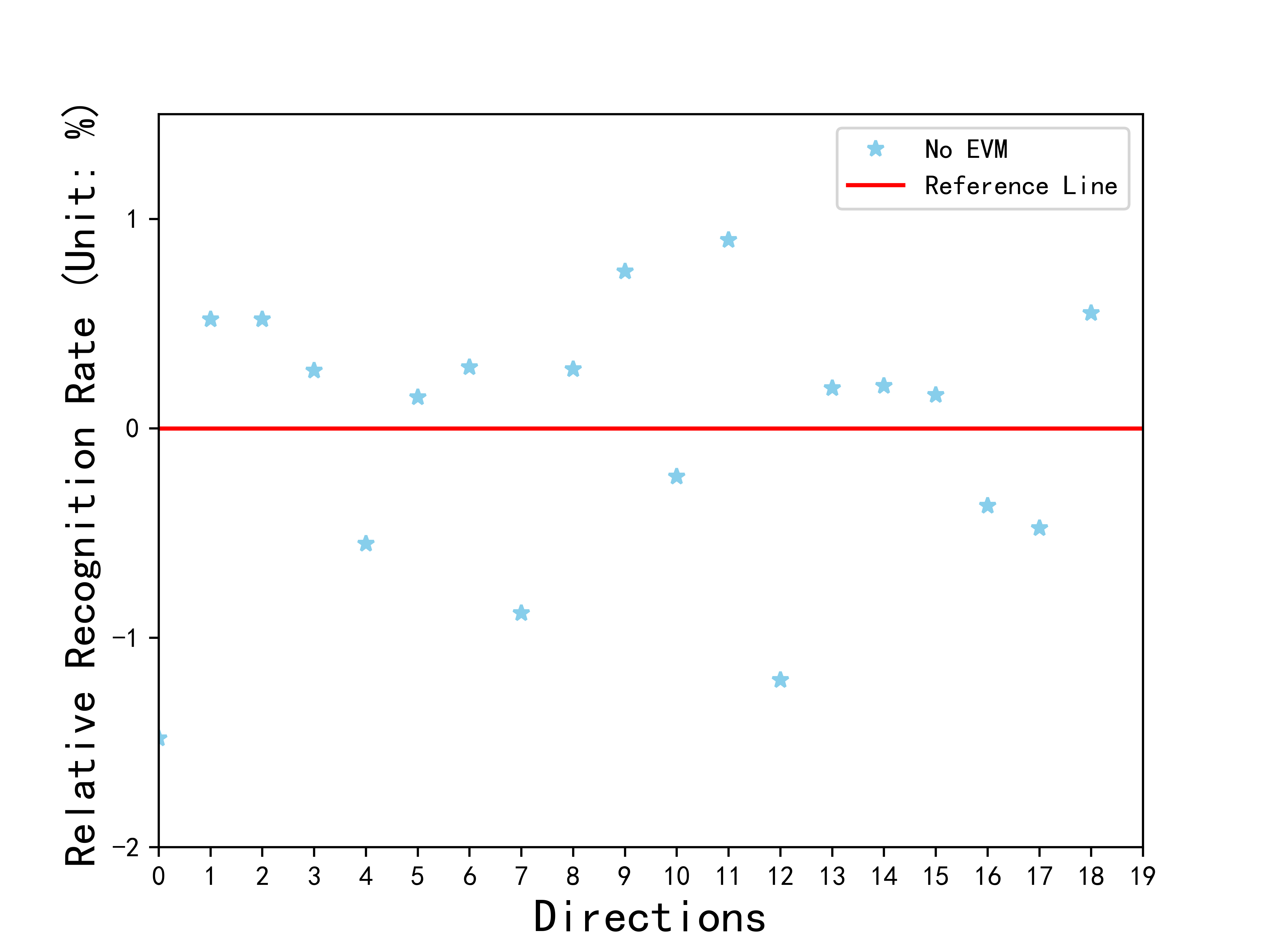}
}

\subfigure [The recognition rate under using EVM.]{
\label{CX}
\includegraphics[height=5.5cm,width=7.9cm]{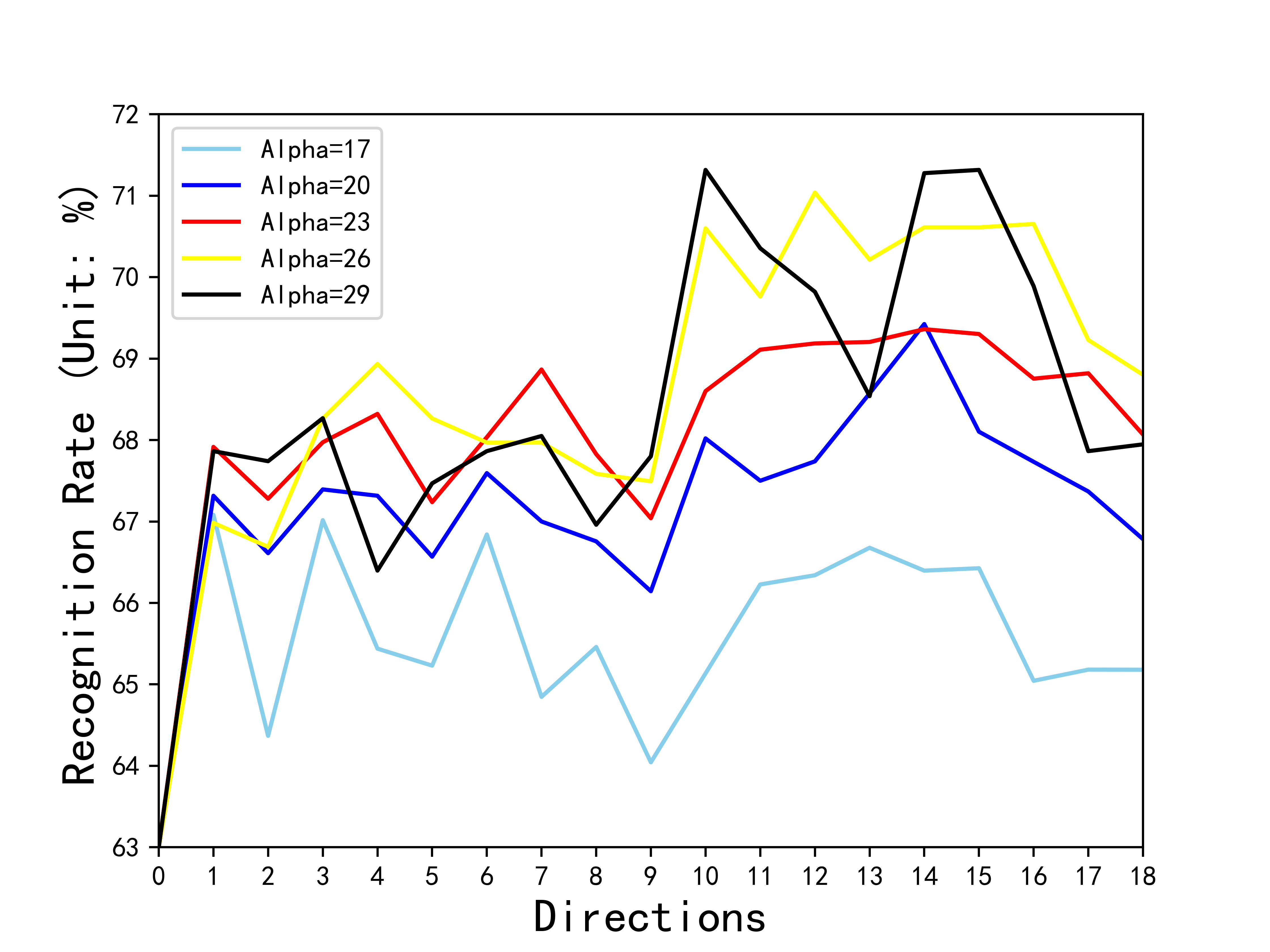}
}
\subfigure [The relative recognition rate under using EVM.]{
\label{CP}
\includegraphics[height=5.5cm,width=7.9cm]{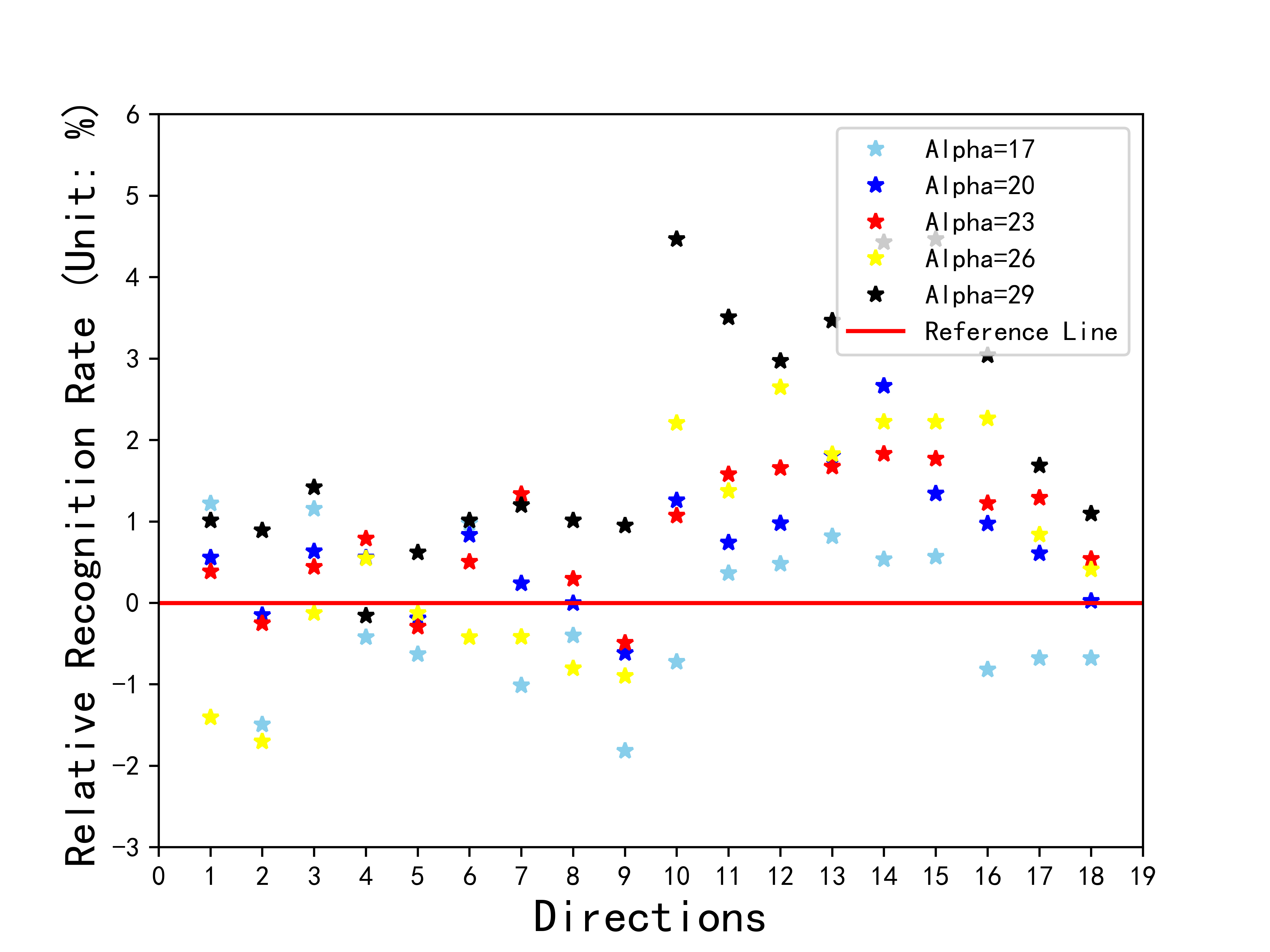}
}
\end{center}
\caption{The CASME \uppercase\expandafter{\romannumeral2} database: the line chart of the recognition rate and the dot charts of relative recognition rate in 18 directions. }
\label{CF}
\end{figure*}%pPp

Our experimental aim is to study movement feature in different directions. In order to analyze the studied aims in detail and comprehensively, this experiment was carried out under two settings: using EVM and not using EVM.

The experiment results not using EVM are reported and analyzed in this paragraph. Table \ref{No} (see appendices) shows the recognition rate of LBP-SDG in 18 directions. The maximum recognition rate (58.57\% on SMIC, 51.48\% on CASME \uppercase\expandafter{\romannumeral2}) of LBP-TOP was used as a reference line. In Figure \ref{SnoX} and \ref{CnoX}, the 18 points (recognition rates in 18 directions) are connected to form a line chart; in Figure \ref{SnoP} and \ref{CnoP}, the relative recognition rate is displayed as 18 independent points. As shown in the two figures,  the movement feature (in directions \circled{1}, \circled{2} and \circled{6} on SMIC-HS, and \circled{1}\circled{2}\circled{9}\circled{11}\circled{18} on CASME \uppercase\expandafter{\romannumeral2}) is very effective; the movement feature (in directions\circled{3}, \circled{8}, \circled{11}, \circled{13} and \circled{16} on SMIC-HS, and \circled{3}, \circled{5}, \circled{6}, \circled{8}, \circled{13}, \circled{14} and \circled{15} on CASME \uppercase\expandafter{\romannumeral2}) is suboptimal; and the movement feature in other directions not only cannot provide discriminative feature information but also produces disruptive feature information.

%PpPppp

The experimental results using EVM are reported and analyzed in this paragraph. Table \ref{S} and \ref{C} (see appendices) show the recognition rates of LBP-SDG in 18 directions under different $Alpha$ values. As shown in Figure \ref{SX} and \ref{CX}, under different $Alpha$ values, the changing trend with the change of direction is consistent in the line chart, and in Figure \ref{SX}, the recognition rate in directions \circled{1}, \circled{3}, \circled{6}, \circled{12}and \circled{16} reaches the local peak, and in direction \circled{2}, \circled{4}, \circled{14} and \circled{17}, it reaches the local trough, which means that the effective directions also tend to be consistent under different $Alpha$ values.  As EVM under different $Alpha$ values can affect the movement features, the effective directions under different $Alpha$ values have somewhat instability, but the overall trends to be consistent. As shown in Figure \ref{SP} and \ref{CP}, with the red line as a reference(the recognition rate of LBP-TOP as shown in Table \ref{DBFeature}), the movement features in direction \circled{1} on SMIC-HS and \circled{11}-\circled{15} on CASME \uppercase\expandafter{\romannumeral2} are very effective, the one in directions \circled{3}, \circled{6}, \circled{7} and \circled{16} on SMIC-HS, and \circled{1}, \circled{3}, \circled{6}, \circled{10}, \circled{16} and \circled{17} on CASME \uppercase\expandafter{\romannumeral2} is suboptimal, and the one in other directions is invalid and even disruptive.

Overall, the studied results on different databases are different. The difference is mainly due to the differences between different databases, such as the country from which the participants came, the collection equipment and environment, etc. These differences and different parameters lead to different effective directions for the two databases and prompted the most suitable $Alpha$ value to be different. Thus, on SMIC-HS, LBP-SDG in directions \circled{1}, \circled{3}, \circled{6} and \circled{16} outperforms LBP-TOP whether EVM is employed or not, and the effective directions are \circled{1}, \circled{3}, \circled{6} and \circled{16}. On the CASME \uppercase\expandafter{\romannumeral2} database, the studied results were divided into two parts: not using EVM, the movement feature in directions \circled{1}, \circled{2}, \circled{9}, \circled{11} and \circled{18} is very effective, and using EVM, the one in directions \circled{11}-\circled{15} is effective. Furthermore, the optimal direction is direction \circled{1} on SMIC-HS and around \circled{14} on CASME \uppercase\expandafter{\romannumeral2}, which means: on SMIC-HS, the variations of the pixel values along the time axis are optimal for distinguishing micro-expressions; on CASME \uppercase\expandafter{\romannumeral2}, the variations of the pixel values in the upward or horizontal direction are very effective for distinguishing the micro-expression.

\subsection{Comparison with Other Features}%PPppppp

\begin{table*}
\setlength{\belowcaptionskip}{8pt}
\renewcommand\tabcolsep{4.0pt}
\caption{Comparison LBP-SDG with other features.}
\centering
\label{DBFeature}
\subtable[On the SMIC-HS]{
\begin{tabular}{ccccccc}
\hline
$Alpha$ & LVP-TOP & GD-LVP-TOP & LBP-SIP & SIP-SDG & LBP-TOP &LBP-SDG  \\
\hline

8  &52.88\% &61.15\% &60.59\% &65.71\%\circled{1} &64.80\% &\textbf{69.68\%}\circled{1}\\
9  &53.65\% &59.60\% &59.69\% &64.01\%\circled{1} &66.81\% &\textbf{69.04\%}\circled{1}\\
10 &51.85\% &59.58\% &59.74\% &64.79\%\circled{1} &64.97\% &\textbf{67.86\%}\circled{1}\\
11 &55.15\% &60.36\% &60.01\% &67.17\%\circled{1} &65.35\% &\textbf{68.67\%}\circled{1}\\
12 &55.04\% &59.09\% &58.54\% &66.62\%\circled{1} &64.82\% &\textbf{69.04\%}\circled{1}\\
\hline
\end{tabular}
       \label{DBFS}
}
\qquad
\subtable[On the CASME \uppercase\expandafter{\romannumeral2}]{
\begin{tabular}{ccccccc}
\hline
$Alpha$ &LVP-TOP &GD-LVP-TOP &LBP-SIP&SIP-SDG& LBP-TOP &LBP-SDG  \\
\hline

17 &62.53\% &65.76\%&58.93\% &62.73\%\circled{15} &65.86\% &\textbf{67.08\%}\circled{1}\\
20 &62.27\% &66.78\%&61.55\% &64.08\%\circled{15} &66.76\% &\textbf{69.42\%}\circled{14}\\
23 &65.48\% &66.41\%&62.92\% &64.28\%\circled{15} &67.53\% &\textbf{69.36\%}\circled{14}\\
26 &64.90\% &66.20\%&63.50\% &65.32\%\circled{15} &68.39\% &\textbf{71.04\%}\circled{12}\\
29 &65.24\% &66.37\%&64.06\% &67.61\%\circled{10} &66.68\% &\textbf{71.32\%}\circled{15}/\circled{10}\\
\hline
\end{tabular}
\label{DBFC}
}
\end{table*}

%pppppp

This experimental aim is to prove the effectiveness of LBP-SDG and to prove that HSDG can provide discriminative feature information. All experiments used EVM, and the results of LBP-SDG are in optimal directions. We compared LBP-SDG with other features, including LVP-TOP, GDLBP-LVP-TOP, LBP-SIP, SIP-SDG, LBP-TOP. LVP and GDLBP-LVP are the LBP variants and were used effectively in face recognition. LVP-TOP and GDLBP-LVP-TOP can be obtained by extracting LVP\cite{B48} and GDLBP-LVP\cite{B47} on three orthogonal planes for micro-expression recognition, and SIP-SDG can be obtained by concatenating LBP-SIP with HSDG. Table \ref{DBFeature} shows the comparative results on both databases, and Table \ref{DBODimension} shows the feature dimension of different features.

Overall, in comparison with other features, LBP-SDG has the best performance under every $Alpha$ value. In detail, SIP-SDG and LBP-SDG have better performance than LBP-SIP and LBP-TOP, respectively, which demonstrates that HSDG provides discriminative feature information. As the performance of SIP-SDG is similar to that of LBP-TOP, and the feature dimension of SIP-SDG is lower than that of LBP-TOP, SIP-SDG can be taken as an alternative features of LBP-TOP; GDLBP-LVP-TOP concatenate LVP-TOP with three LBP-TOP, but in comparison with LBP-TOP, both LVP-TOP and GDLBP-LVP-TOP have worse performance, which demonstrates the following: 1) the effective features in other fields may not be effective for micro-expression recognition, and may even worsen the recognition rate; and 2) the proposed HSDG not only has a low feature dimension but also can enhance the performance of LBP-TOP/LBP-SIP, which shows that HSDG is a discriminative feature to represent micro-expressions. Furthermore, we compare our method with HOG-TOP and HIGO-TOP using EVM, and the results of the two features can be found in \cite{B19}. As shown in Table \ref{DBOH}, LBP-SDG is superior to HOG-TOP and HIGO-TOP. In terms of the highest recognition rate, LBP-SDG outperforms these features on both databases.
%for the SMIC database, LBP-SDG is 14.53\%, 8.53\%, 9.09\%, 2.51\%, 2.87\% 8.09\% and 1.39\% higher than LVP-TOP, GDLBP-LVP-TOP, LBP-SIP, SIP-SDG, LBP-TOP, HOG-TOP and HIGO-TOP, respectively; on CASME \uppercase\expandafter{\romannumeral2} database, the corresponding values are 5.84\%, 4.54\%, 7.26\%, 3.71\%, 2.93\%, 7.35\% and 4.11\%.

We further compare the confusion matrices of LVP-TOP, GDLBP-LVP-TOP, LBP-SIP, SIP-SDG, LBP-TOP and LBP-SDG in Figure \ref{SConF} and \ref{CConF} (see appendices). On SMIC, LBP-SDG does not have the best performance at recognizing a certain class of expression, while it has the best average performance over three classes. On CASME \uppercase\expandafter{\romannumeral2}, SIP-SDG has the best performance at recognizing ``Happiness"; GDLBP-LVP-TOP has best performance at recognizing ``Surprise"; LVP-TOP, LBP-SIP and LBP-TOP has best performance for recognizing ``Others"; for the other two micro-expressions(``Disgust", ``Repression"), all feature have poor performance, while LBP-SDG has the best performance (57\% for ``Disgust" and 56\% for ``Repression"); In terms of the average performance, LBP-SDG is optimal. Furthermore, according to Table \ref{DBODimension}, HSDG has a very small feature dimension, which makes the feature dimension of LBP-SDG (SIP-SDG) very close to that of LBP-TOP (LBP-SIP); LVP-TOP and GDLBP-LVP-TOP have a very high feature dimension, which seriously reduces the speed of recognition.

\begin{table}%p
\large
\setlength{\belowcaptionskip}{8pt}
\begin{center}
\caption{Comparison of feature dimension.}
\centering
\label{DBODimension}
\begin{tabular}{ccc}
\hline
Methods &CASME \uppercase\expandafter{\romannumeral2} & SMIC-HS  \\
\hline
HSDG& 50  &256 \\
LVP-TOP & 17700  & 90624\\
GD-LVP-TOP &30975 &109056 \\
LBP-SIP &1500 &7680 \\
SIP-SDG &1550 &7930 \\
LBP-TOP &4425 &6144 \\
LBP-SDG &4475 &6400 \\
\hline
\end{tabular}
\end{center}

\end{table}
%p

\subsection{Comparison with Other Methods}%ppppppp

This section proves that EVM+LBP-SDG is competitive. The recognition rate of different methods is shown in Table \ref{DBOM}. it can be seen that EVM+LBP-SDG has the best performance (69.68\% on SMIC-HS and 71.32\% on CASME \uppercase\expandafter{\romannumeral2}). CEF is the state-of-the-art method in the traditional methods. Although CEF employs an additional model to integrate some processing, EVM+LBP-SDG still has better performance. LEARNet, SSSN and TSCNN are the state-of-the-art methods in deep learning-based methods. First, as shown in Table \ref{DBOM}, LEARNet has the highest recognition rate, while it adopts N-fold cross-validation and not the mainstream cross-validation method (LOSO); second, comparing SSSN, EVM+LBP-SDG shows only a slight advantage on CASME \uppercase\expandafter{\romannumeral2}, but on SMIC-HS, EVM+LBP-SDG has an obvious advantage; third, in terms of recognition rate, TSCNN has the best results, but similar to SSSN, it needs to detect apex frame before recognizing micro-expressions, and in terms of inputs, TSCNN need to calculate optical flow, while our method does not; finally, the deep learning-based methods train a feature extractor and the classifier for each subject validation, while the traditional methods (including our method) uses the same feature extractor for all subject validation.

\begin{table}
\large
\setlength{\belowcaptionskip}{8pt}
\begin{center}
\caption{Comparison of accuracy rate between LBP-SDG, HOG-TOP and HIGO-TOP.}
\centering
\label{DBOH}
\begin{tabular}{ccc}
\hline
Methods & CASME \uppercase\expandafter{\romannumeral2} & SMIC-HS  \\
\hline
HOG-TOP & 63.97  &61.59 \\
HIGO-TOP & 67.21  &68.29\\
LBP-SDG &\textbf{71.32 } &\textbf{69.68} \\
\hline
\end{tabular}
\end{center}

\end{table}

\begin{table}
\begin{center}
\caption{Comparison of recognition rate between the proposed method and other methods.}
\label{DBOM}
\setlength{\tabcolsep}{4pt}

\begin{threeparttable}

\begin{tabular}{ccc}
\hline
Method & SMIC-HS & CASME \uppercase\expandafter{\romannumeral2} \\
\hline
FDM~\cite{B21} & 54.88\% &45.93\%\\
STCLQP\cite{B12}  &64.02\% &58.39\%\\
STLBP-IP~\cite{B13} &57.93\% &59.51\%\\
TIM+STRBP~\cite{B45} &60.98\% &64.37\%\\
CNN+LSTM~\cite{B27}  & - &60.98\%\\
ELRCN-TE~\cite{B46}  &- &52.44\%\\
STLBP-IP+KGSL~\cite{B15} &60.78\% &63.83\%\\
3DFCNN~\cite{B29}    &55.49\% &59.11\%\\
TIM+EVM+HIGO~\cite{B19} &68.29\% &67.21\%\\
LEARNet*~\cite{B44}  &- &76.59\%\\
SSSN$^\curlyvee$~\cite{B43}   &63.41 &71.19\%\\
TSCNN$^\curlyvee$~\cite{B59} &72.74 &74.05\%\\
CEF~\cite{B17}&68.90\% &70.85\%\\
The proposed method & \textbf{69.68\%}&\textbf{71.32\%}\\
\hline
\end{tabular}
\begin{tablenotes}
        \footnotesize
        \item[-] no result in the original paper  %此处加入注释*信息
        \item[$^\curlyvee$] the method uses the apex frame %此处加入注释**信息
        \item[*] the results adopted different cross-validation. %此处加入注释**信息
      \end{tablenotes}
\end{threeparttable}
\end{center}
\end{table}%p

\section{Future Works}%ppppPpppppp
In this work, HSDGs in 18 directions were tested in turn to artificially select effective HSDGs, which spent a lot of time and didn't consider the multiple HSDGs simultaneously. In addition, different parameters and databases will lead to different conclusions about the effective HSDGs. In future works, we will solve it by employing the feature selection algorithm that can automatically select effective HSDGs and process 18 HSDGs simultaneously. According to the work in this paper, LBP-TOP and HSDG have different importance. Concretely, LBP-TOP is the main feature that provides abundant feature information, and HSDGs is the supplementary feature that provides the limited feature information and different HSDGs have different performance. Thus, this algorithm selects effective HSDGs under concatenating with LBP-TOP and should consider the different importance of LBP-TOP and HSDG. In this way, we needn't test each HSDG in turn and can consider multiple HSDGs simultaneously. In addition, more discriminative features can be selected to recognize micro-expressions, and the performance can further be improved.

\section{Conclusions}%ppppPpp
This paper studies movement feature in different directions and proposes a new and low-dimensional single-direction movement feature HSDG. HSDG actively extracts the movement feature in a certain direction. First, the gradient values are calculated in a certain direction, then these values are quantified before performing histogram operations.  Concatenating HSDG with LBP-TOP, we obtain LBP-SDG. In experiments, LBP-SDG in 18 directions was tested to ensure effective and optimal directions, and LBP-SDG in an optimal direction was compared with other features and methods. On SMIC-HS, the movement features along the time axis are optimal for distinguishing micro-expressions; on CASME \uppercase\expandafter{\romannumeral2}, extracting the movement features in the upward or horizontal directions is very effective for distinguishing the micro-expression. Comparing it with other features, LBP-SDG in an optimal direction has the best performance. By comparing the changes of LBP-TOP and LBP-SIP before and after adding HSDG, it is proven that HSDG can provide discriminative feature information. Additionally, the results show that LBP-SDG+EVM outperforms state-of-the-art methods.

\section{Acknowledgements}%
This work was partly supported by the Postgraduate Research and Practice Innovation Program of Jiangsu Province (Grant KYCX18$\_$0899), partly by the National Natural Science Foundation of China (NSFC) under Grants 72074038, partly by the Key Research and Development Program of Jiangsu Province(Grant BE2016775), partly by the National Natural Science Foundation of China (NSFC) (Grants 61971236), partly by China Postdoctoral Science Foundation (Grant 2018M632348).

\bibliography{ms}
%\appendixpp
\section*{Appendices}

\begin{table*}[htb]
\setlength{\belowcaptionskip}{8pt}
\renewcommand\tabcolsep{4.0pt}
\caption{The recognition rate of LBP-SDG in 18 directions under not using EVM.}
\centering
\label{No}
\subtable[On the SMIC-HS]{
\begin{tabular}{cccc}
\hline
DT&RR($R_V$,$R$)&DT&RR($R_V$,$R$)\\
\hline
\circled{1}&\textbf{63.15\%}(1,2)&\circled{10}&58.55\%(1,4)\\
\circled{2} &61.85\%(3,3) &\circled{11}&60.88\%(3,3)\\
\circled{3} &59.61\%(1,1) &\circled{12}&58.38\%(1,1)\\
\circled{4} &57.07\%(1,1) &\circled{13}&59.09\%(3,1)\\
\circled{5} &57.42\%(2,1) &\circled{14}&57.68\%(1,1)\\
\circled{6} &62.00\%(1,1) &\circled{15}&58.23\%(1,1)\\
\circled{7} &56.93\%(1,1) &\circled{16}&59.51\%(1,2)\\
\circled{8} &58.80\%(2,1) &\circled{17}&58.50\%(2,4)\\
\circled{9} &57.38\%(1,2) &\circled{18}&58.10\%(1,2)\\
\hline
\end{tabular}
       \label{Sno}
}
\qquad
\subtable[On the CASME \uppercase\expandafter{\romannumeral2}]{
\begin{tabular}{cccc}
\hline
DT&RR($R_V$,$R$)&DT&RR($R_V$,$R$)\\
\hline
\circled{1}&52.00\%(5,3-4)&\circled{10}&51.88\%(5,4)\\
\circled{2} &52.00\%(5,2-3 5) &\circled{11}&\textbf{52.38\%}(8,5)\\
\circled{3} &51.79\%(4,8) &\circled{12}&50.48\%(4,8)\\
\circled{4} &50.93\%(3,2-3) &\circled{13}&52.21\%(4,7)\\
\circled{5} &51.63\%(5,7) &\circled{14}&51.71\%(5,7)\\
\circled{6} &51.77\%(5,3) &\circled{15}&51.65\%(5,8)\\
\circled{7} &50.60\%(4,6-8) &\circled{16}&51.11\%(7,3-7)\\
\circled{8} &51.76\%(4,8) &\circled{17}&51.00\%(4,2-3)\\
\circled{9} &52.23\%(5,2) &\circled{18}&52.03\%(5,4-6)\\
\hline
\end{tabular}
\label{Cno}
}
\end{table*}

\begin{table*}%%p
\large
\setlength{\belowcaptionskip}{8pt}
\caption{The SMIC-HS database: the recognition rate of LBP-SDG in 18 directions under different $Alpha$ values.}
\centering
\label{S}
\setlength{\tabcolsep}{4pt}
\subtable[$Alpha$=8]{
\begin{tabular}{cccc}
\hline
DT&RR($R_V$,$R$)&DT&RR($R_V$,$R$)\\
\hline
\circled{1}&\textbf{69.68\%}(3,3)&\circled{10}&63.36\%(3,3)\\
\circled{2} &65.10\%(3,3) &\circled{11}&65.05\%(3,4)\\
\circled{3} &67.57\%(3,2) &\circled{12}&67.72\%(3,1)\\
\circled{4} &63.32\%(3,1) &\circled{13}&61.98\%(3,1)\\
\circled{5} &63.72\%(3,1) &\circled{14}&62.69\%(3,1)\\
\circled{6} &65.67\%(3,2) &\circled{15}&63.88\%(3,2)\\
\circled{7} &65.02\%(3,1) &\circled{16}&65.64\%(3,2)\\
\circled{8} &65.86\%(3,2) &\circled{17}&65.38\%(3,1)\\
\circled{9} &63.12\%(3,2) &\circled{18}&65.64\%(3,1)\\
\hline
\end{tabular}
       \label{alpha8S}
}
\qquad
\subtable[$Alpha$=9]{\begin{tabular}{cccc}
\hline
DT&RR($R_V$,$R$)&DT&RR($R_V$,$R$)\\
\hline
\circled{1}&\textbf{69.04\%}(3,3)&\circled{10}&64.71\%(3,2)\\
\circled{2} &66.01\%(3,1) &\circled{11}&64.08\%(3,2)\\
\circled{3} &66.83\%(3,3) &\circled{12}&66.22\%(3,1)\\
\circled{4} &64.19\%(3,1) &\circled{13}&63.94\%(3,4)\\
\circled{5} &66.05\%(3,1) &\circled{14}&63.17\%(3,2)\\
\circled{6} &66.24\%(3,2) &\circled{15}&65.86\%(3,1)\\
\circled{7} &64.39\%(3,1) &\circled{16}&64.63\%(3,2)\\
\circled{8} &65.51\%(3,2) &\circled{17}&64.30\%(3,2)\\
\circled{9} &66.27\%(3,2) &\circled{18}&65.12\%(3,1)\\
\hline
\end{tabular}
       \label{alpha9S}
}
\qquad
\subtable[$Alpha$=10]{\begin{tabular}{cccc}
\hline
DT&RR($R_V$,$R$)&DT&RR($R_V$,$R$)\\
\hline
\circled{1}&\textbf{69.00\%}(3,3)&\circled{10}&62.30\%(3,4)\\
\circled{2} &64.64\%(3,3) &\circled{11}&64.46\%(3,4)\\
\circled{3} &66.89\%(3,2) &\circled{12}&65.10\%(3,1)\\
\circled{4} &63.93\%(4,1) &\circled{13}&63.90\%(3,3)\\
\circled{5} &64.23\%(3,1) &\circled{14}&63.08\%(3,1)\\
\circled{6} &68.37\%(3,3) &\circled{15}&63.08\%(3,1)\\
\circled{7} &66.11\%(3,1) &\circled{16}&67.35\%(3,3)\\
\circled{8} &65.61\%(3,2) &\circled{17}&63.58\%(4,1)\\
\circled{9} &64.23\%(3,2) &\circled{18}&64.19\%(3,1)\\
\hline
\end{tabular}
       \label{alpha10S}
}
\qquad
\subtable[$Alpha$=11]{\begin{tabular}{cccc}
\hline
DT&RR($R_V$,$R$)&DT&RR($R_V$,$R$)\\
\hline
\circled{1}&\textbf{68.67\%}(3,3)&\circled{10}&62.66\%(3,4)\\
\circled{2} &64.84\%(3,2) &\circled{11}&63.65\%(3,2)\\
\circled{3} &67.32\%(3,1) &\circled{12}&64.22\%(3,2)\\
\circled{4} &64.89\%(4,1) &\circled{13}&63.29\%(3,3)\\
\circled{5} &64.43\%(3,1) &\circled{14}&62.47\%(4,1)\\
\circled{6} &65.78\%(3,2) &\circled{15}&63.53\%(3,1)\\
\circled{7} &66.03\%(3,1) &\circled{16}&67.27\%(3,2)\\
\circled{8} &63.62\%(3,3) &\circled{17}&64.04\%(3,3)\\
\circled{9} &63.93\%(3,2) &\circled{18}&65.01\%(3,1)\\

\hline
\end{tabular}
       \label{alpha11S}
}
\qquad
\subtable[$Alpha$=12]{\begin{tabular}{cccc}
\hline
DT&RR($R_V$,$R$)&DT&RR($R_V$,$R$)\\
\hline
\circled{1}&\textbf{69.04\%}(3,3)&\circled{10}&63.41\%\\
\circled{2} &64.69\%(4,4) &\circled{11}&63.72\%(4,2)\\
\circled{3} &66.68\%(3,2) &\circled{12}&64.05\%(3,1)\\
\circled{4} &63.79\%(4,1) &\circled{13}&62.23\%(3,3)\\
\circled{5} &64.31\%(3,2) &\circled{14}&62.06\%(4,1)\\
\circled{6} &67.05\%(3,3) &\circled{15}&64.15\%(4,1)\\
\circled{7} &65.15\%(4,1) &\circled{16}&65.99\%(3,2)\\
\circled{8} &64.37\%(3,2) &\circled{17}& 63.67\%(3,4)\\
\circled{9} &65.73\%(3,3) &\circled{18}&64.37\%(3,1)\\
\hline
\end{tabular}
       \label{alpha12S}
}
\end{table*}

\begin{table*}%%pP
\large
\setlength{\belowcaptionskip}{8pt}
\caption{The CASME \uppercase\expandafter{\romannumeral2} database: the recognition rate of LBP-SDG in 18 directions under different $Alpha$ values.}
\centering
\label{C}
\setlength{\tabcolsep}{4pt}
\subtable[$Alpha$=17]{
\begin{tabular}{cccc}
\hline
DT&RR($R_V$,$R$)&DT&RR($R_V$,$R$)\\
\hline
\circled{1}&\textbf{67.08\%}(4,8)&\circled{10}&65.14\%(5,4)\\
\circled{2} &64.37\%(4,2 4) &\circled{11}&66.23\%(4,4)\\
\circled{3} &67.02\%(6,8) &\circled{12}&66.34\%(4,8)\\
\circled{4} &65.44\%(5,2) &\circled{13}&66.68\%(4,8)\\
\circled{5} &65.23\%(6,8) &\circled{14}&66.40\%(4,8)\\
\circled{6} &66.84\%(6,8) &\circled{15}&66.43\%(6,8)\\
\circled{7} &64.85\%(5,3) &\circled{16}&65.04\%(5,4)\\
\circled{8} &65.46\%(6,7) &\circled{17}&65.18\%(4,8)\\
\circled{9} &64.04\%(4,5) &\circled{18}&65.18\%(4,3)\\
\hline
\end{tabular}
       \label{alpha17C}
}
\qquad
\subtable[$Alpha$=20]{\begin{tabular}{cccc}
\hline
DT&RR($R_V$,$R$)&DT&RR($R_V$,$R$)\\
\hline
\circled{1}&67.32\%(6,7)&\circled{10}&68.02\%(6,5)\\
\circled{2} &66.61\%(6,3) &\circled{11}&67.50\%(5,5)\\
\circled{3} &67.39\%(5,7) &\circled{12}&67.74\%(4,4)\\
\circled{4} &67.32\%(6,8) &\circled{13}&68.57\%(6,6)\\
\circled{5} &66.57\%(6,8) &\circled{14}&\textbf{69.42\%}(4,8)\\
\circled{6} &67.59\%(5,8) &\circled{15}&68.10\%(4,8)\\
\circled{7} &67.00\%(6,8) &\circled{16}&67.73\%(4,6)\\
\circled{8} &66.76\%(4,4) &\circled{17}&67.37\%(4,4)\\
\circled{9} &66.14\%(4,3) &\circled{18}&66.78\%(8,4)\\
\hline
\end{tabular}
       \label{alpha20C}
}
\qquad
\subtable[$Alpha$=23]{\begin{tabular}{cccc}
\hline
DT&RR($R_V$,$R$)&DT&RR($R_V$,$R$)\\
\hline
\circled{1}&67.92\%(5,2)&\circled{10}&68.60\%(4,6)\\
\circled{2} &67.28\%(8,4) &\circled{11}&69.11\%(7,4)\\
\circled{3} &67.97\%(8,6) &\circled{12}&69.19\%(4,2)\\
\circled{4} &68.32\%(4,2) &\circled{13}&69.20\%(4,6)\\
\circled{5} &67.24\%(7,6) &\circled{14}&\textbf{69.36\%}(5,8)\\
\circled{6} &68.03\%(7,3) &\circled{15}&69.30\%(3,8)\\
\circled{7} &68.87\%(4,2) &\circled{16}&68.75\%(5,5)\\
\circled{8} &67.83\%(5,2) &\circled{17}&68.82\%(6,7)\\
\circled{9} &67.04\%(6,3) &\circled{18}&68.07\%(7,5)\\
\hline
\end{tabular}
       \label{alpha23C}
}
\qquad
\subtable[$Alpha$=26]{\begin{tabular}{cccc}
\hline
DT&RR($R_V$,$R$)&DT&RR($R_V$,$R$)\\
\hline
\circled{1}&66.98\%(5,2)&\circled{10}&70.60\%(6,8)\\
\circled{2} &66.69\%(5,3) &\circled{11}&69.76\%(5,4)\\
\circled{3} &68.27\%(5,3) &\circled{12}&\textbf{71.04\%(5,8)}\\
\circled{4} &68.93\%(5,2) &\circled{13}&70.22\%(6,5)\\
\circled{5} &68.27\%(5,4) &\circled{14}&70.61\%(5,8)\\
\circled{6} &67.97\%(5,2) &\circled{15}&70.61\%(5,8)\\
\circled{7} &67.97\%(5,3) &\circled{16}&70.65\%(6,6)\\
\circled{8} &67.59\%(5,2) &\circled{17}&69.23\%(5,4)\\
\circled{9} &67.49\%(7,2) &\circled{18}&68.80\%(5,5)\\

\hline
\end{tabular}
       \label{alpha26C}
}
\qquad
\subtable[$Alpha$=29]{\begin{tabular}{cccc}
\hline
DT&RR($R_V$,$R$)&DT&RR($R_V$,$R$)\\
\hline
\circled{1}&67.86\%(7,4)&\circled{10}&\textbf{71.32\%}(5,8)\\
\circled{2} &67.74\%(7,2) &\circled{11}&70.36\%(5,8)\\
\circled{3} &68.27\%(7,8) &\circled{12}&69.82\%(6,8)\\
\circled{4} &66.69\%(7,4) &\circled{13}&70.32\%(5,8)\\
\circled{5} &67.47\%(8,4) &\circled{14}&71.28\%(5,8)\\
\circled{6} &67.86\%(7,4) &\circled{15}&\textbf{71.32\%}(5,8)\\
\circled{7} &68.05\%(5,4) &\circled{16}&69.89\%(5,6)\\
\circled{8} &67.86\%(7,2) &\circled{17}&68.54\%(5,8)\\
\circled{9} &67.80\%(5,3) &\circled{18}&67.95\%(5,8)\\
\hline
\end{tabular}
       \label{alpha29C}
}
\end{table*}

\begin{figure*}[htb]
\begin{center}
\subfigure [LVP-TOP.]{
\label{SLVP}
\includegraphics[height=5cm,width=7.8cm]{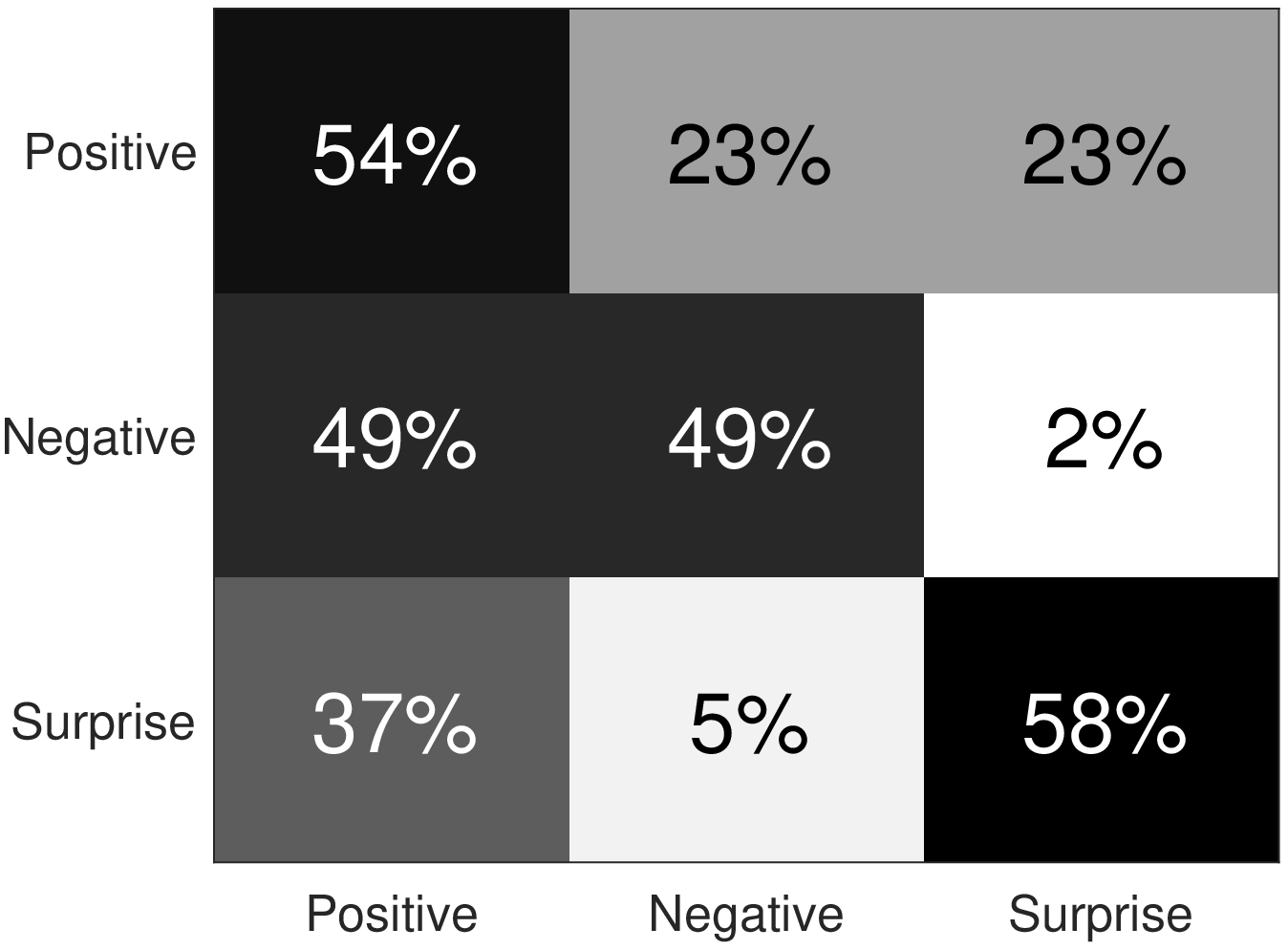}
}
\subfigure [GDLBP-LVP-TOP]{
\label{SGDLVP}
\includegraphics[height=5cm,width=7.8cm]{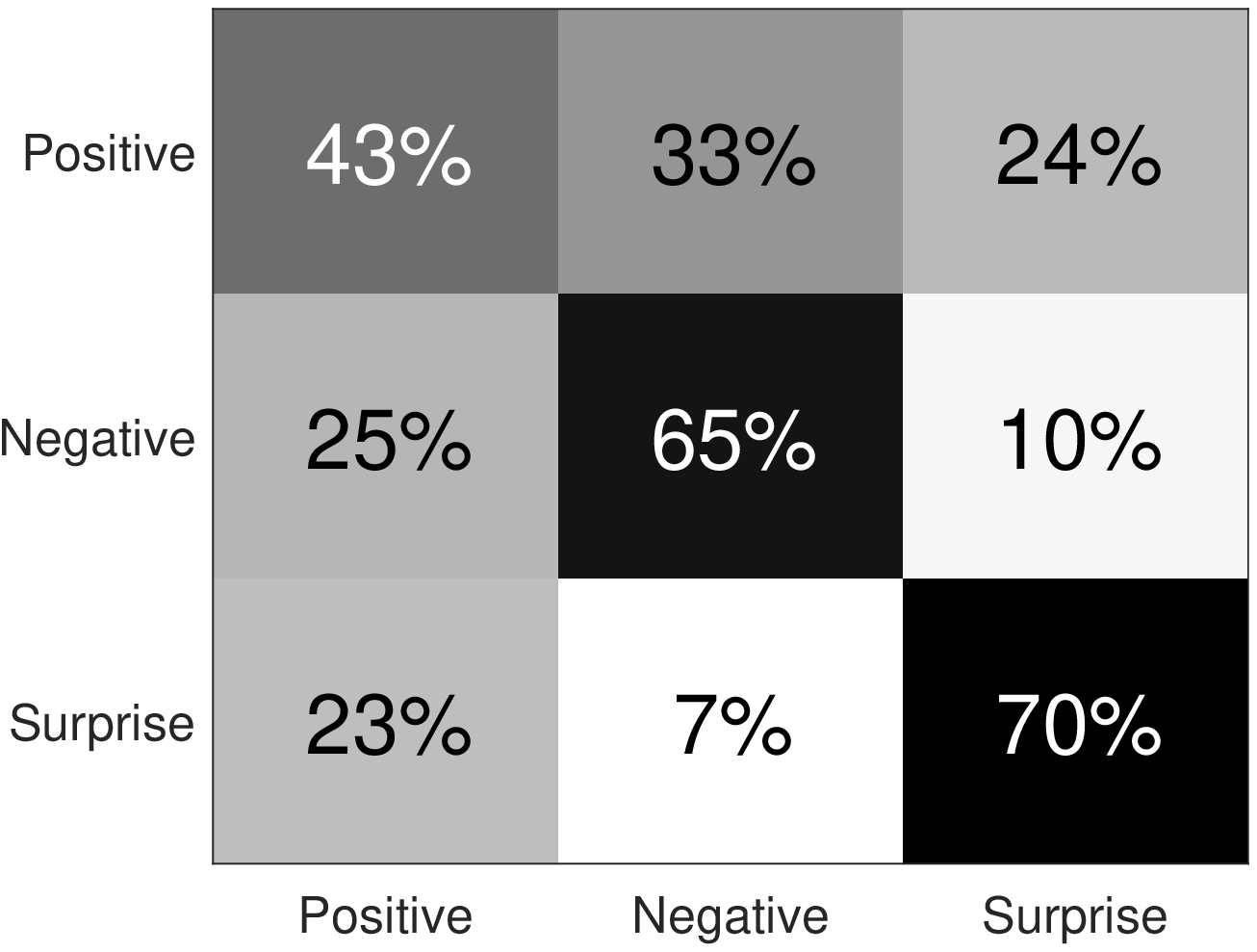}
}
\subfigure [LBP-SIP.]{
\label{SSIP}
\includegraphics[height=5cm,width=7.8cm]{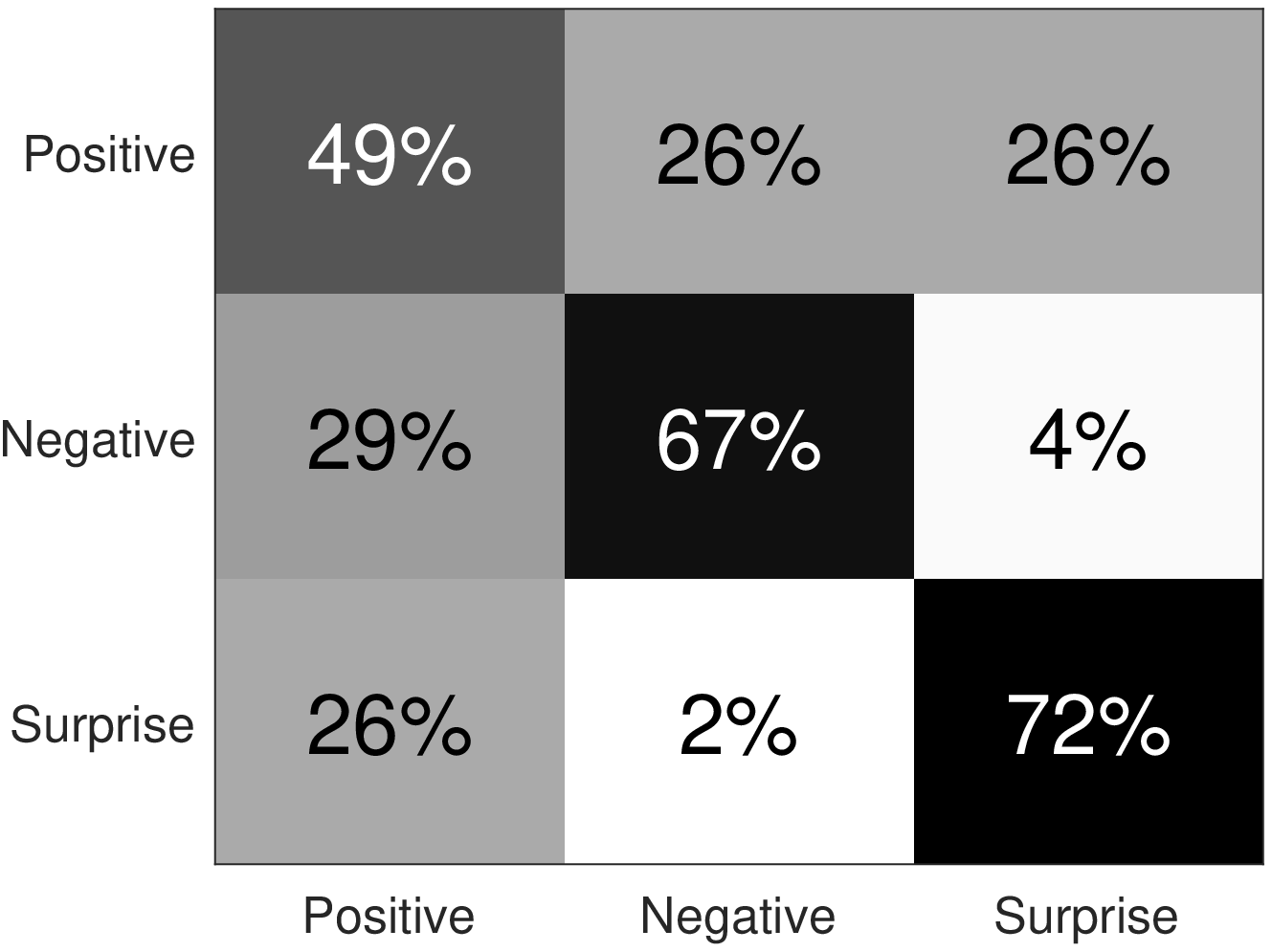}
}
\subfigure [SIP-SDG]{
\label{SSIP-SDG}
\includegraphics[height=5cm,width=7.8cm]{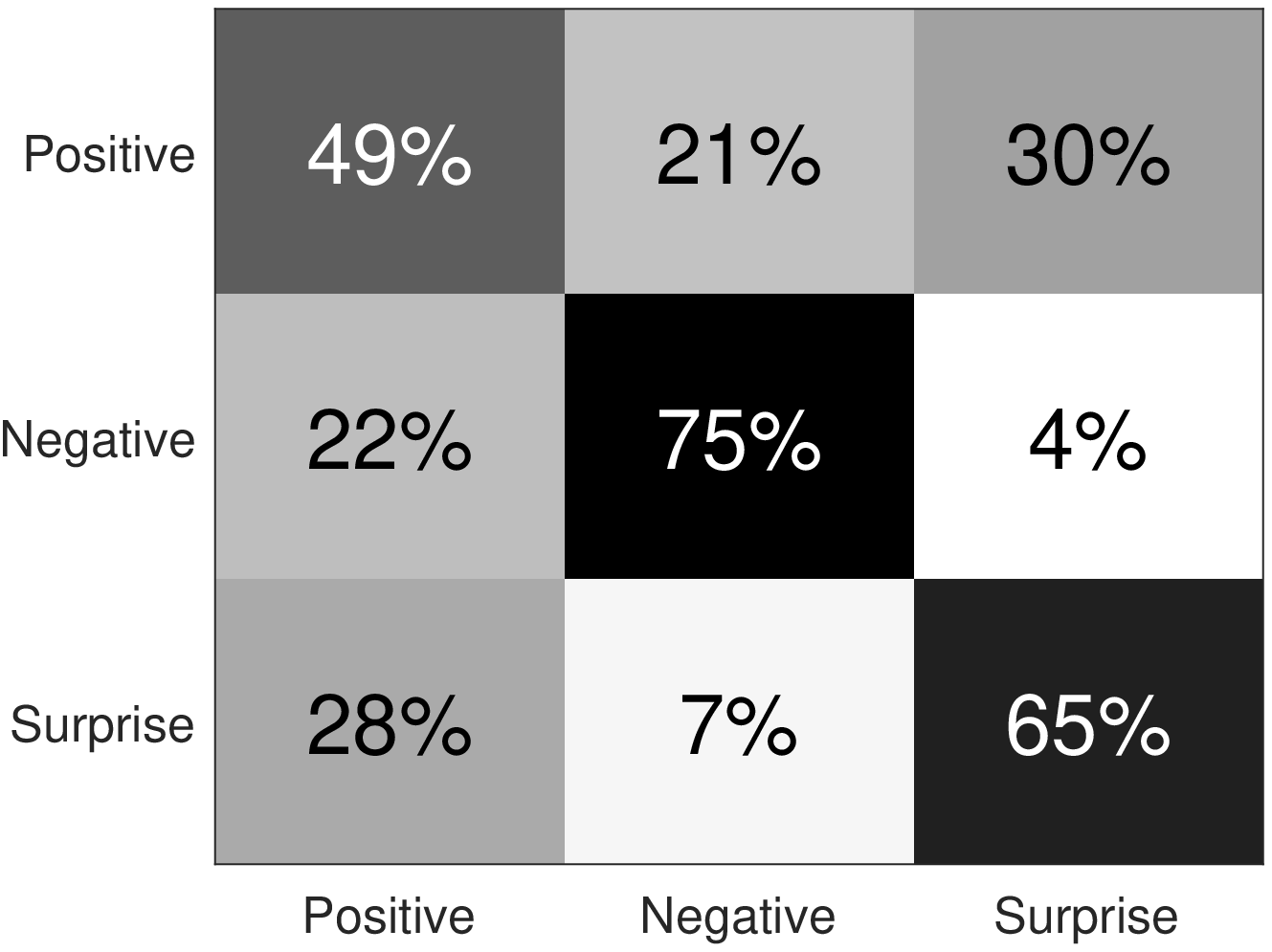}
}C
\subfigure [LBP-TOP.]{
\label{STOP}
\includegraphics[height=5cm,width=7.8cm]{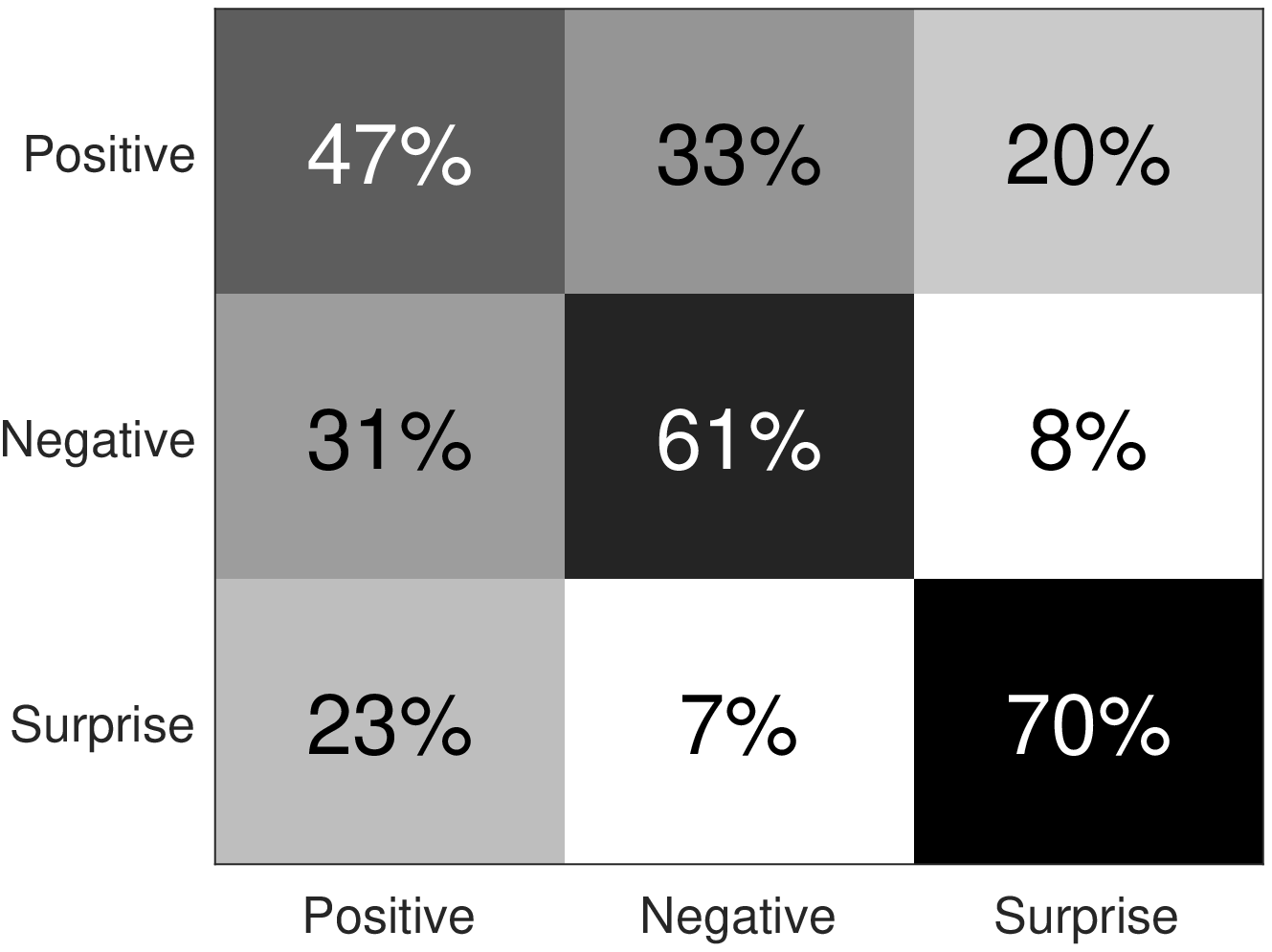}
}
\subfigure [LBP-SDG]{
\label{SSDG}
\includegraphics[height=5cm,width=7.8cm]{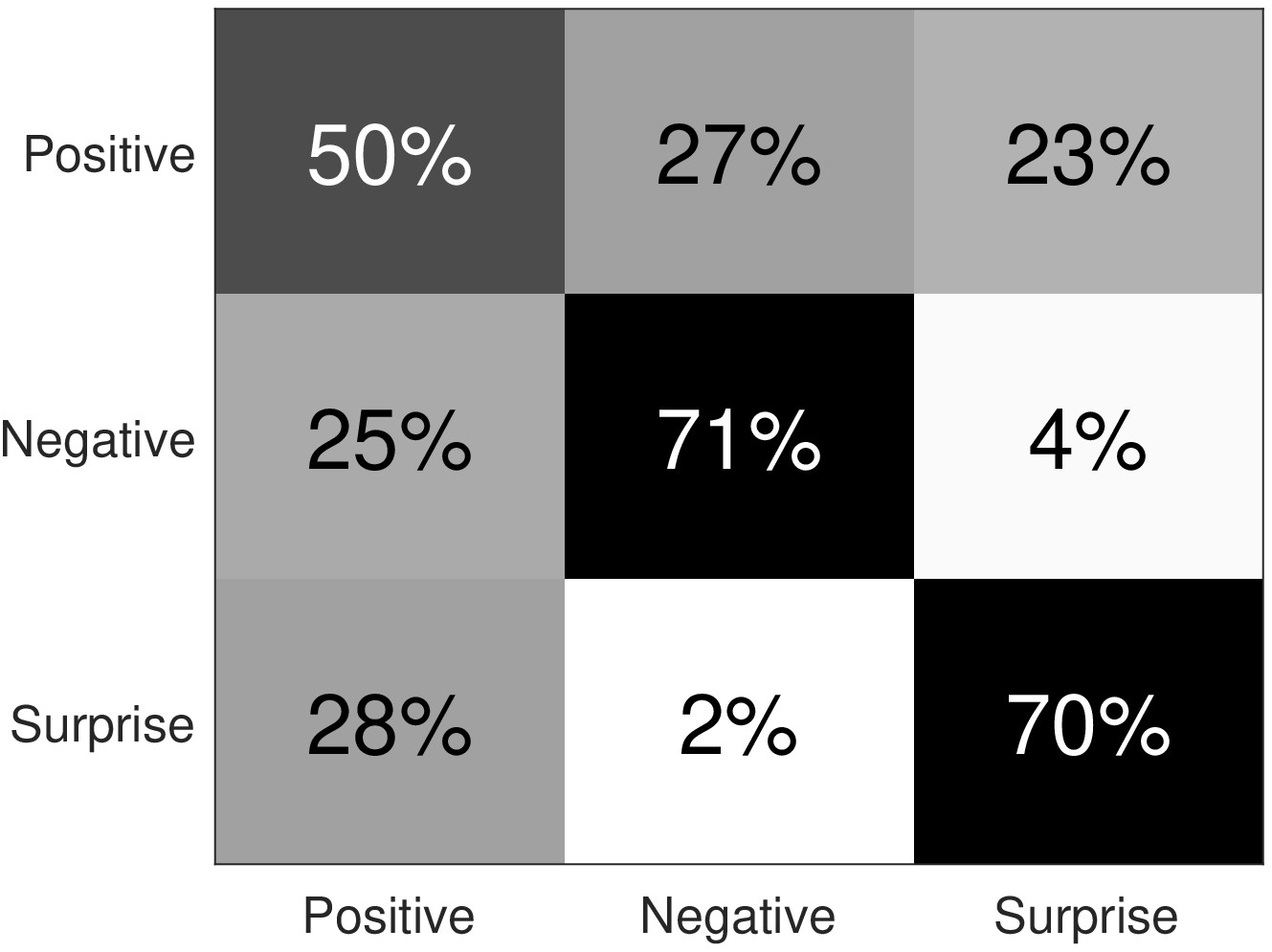}
}
\end{center}
\caption{The confusion matrices of different features on the SMIC-HS database.}
\label{SConF}
\end{figure*}

\begin{figure*}[htb]
\begin{center}
\subfigure [LVP-TOP.]{
\label{CLVP}
\includegraphics[height=5cm,width=7.8cm]{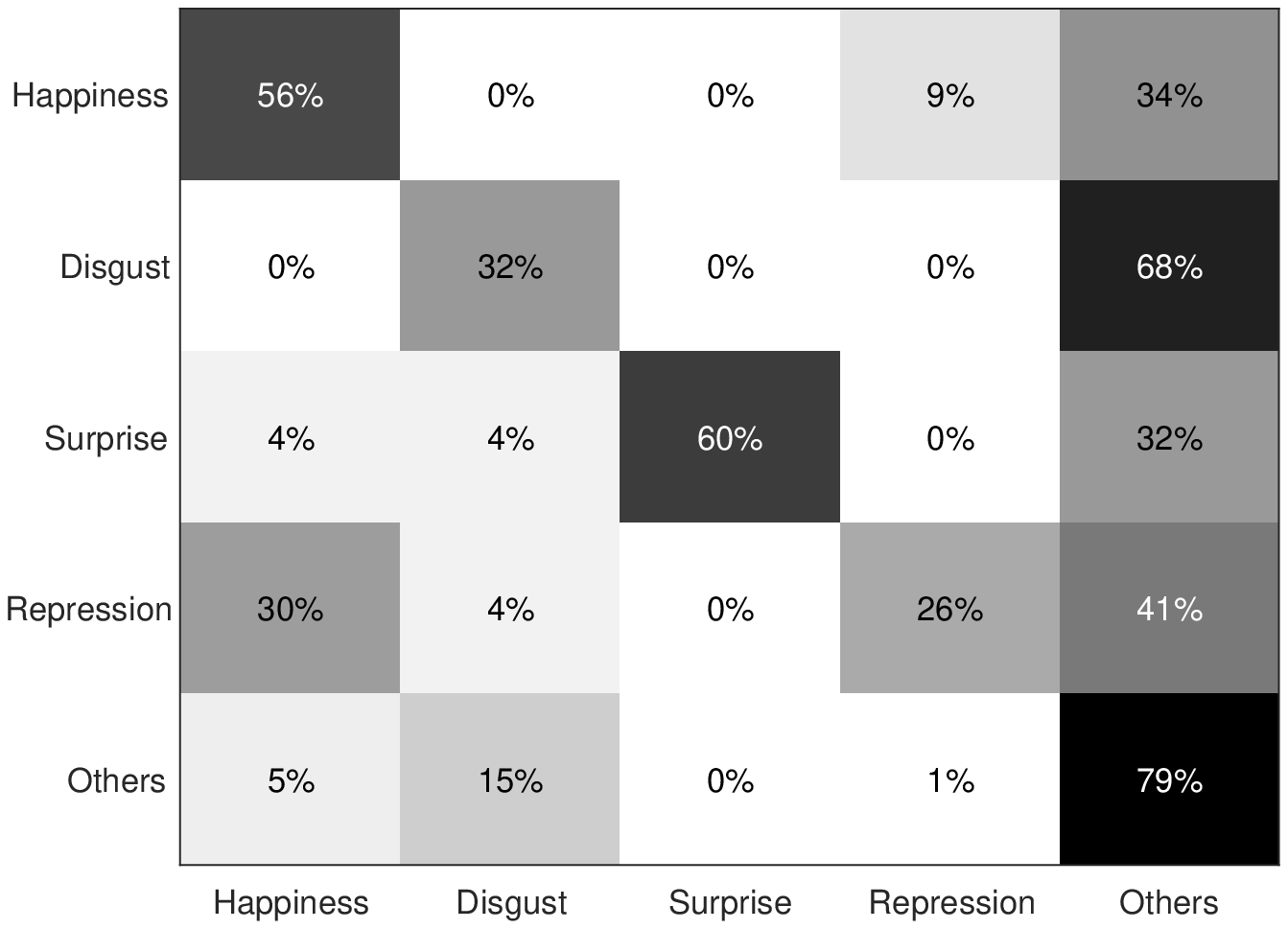}
}
\subfigure [GDLBP-LVP-TOP]{
\label{CGDLVP}
\includegraphics[height=5cm,width=7.8cm]{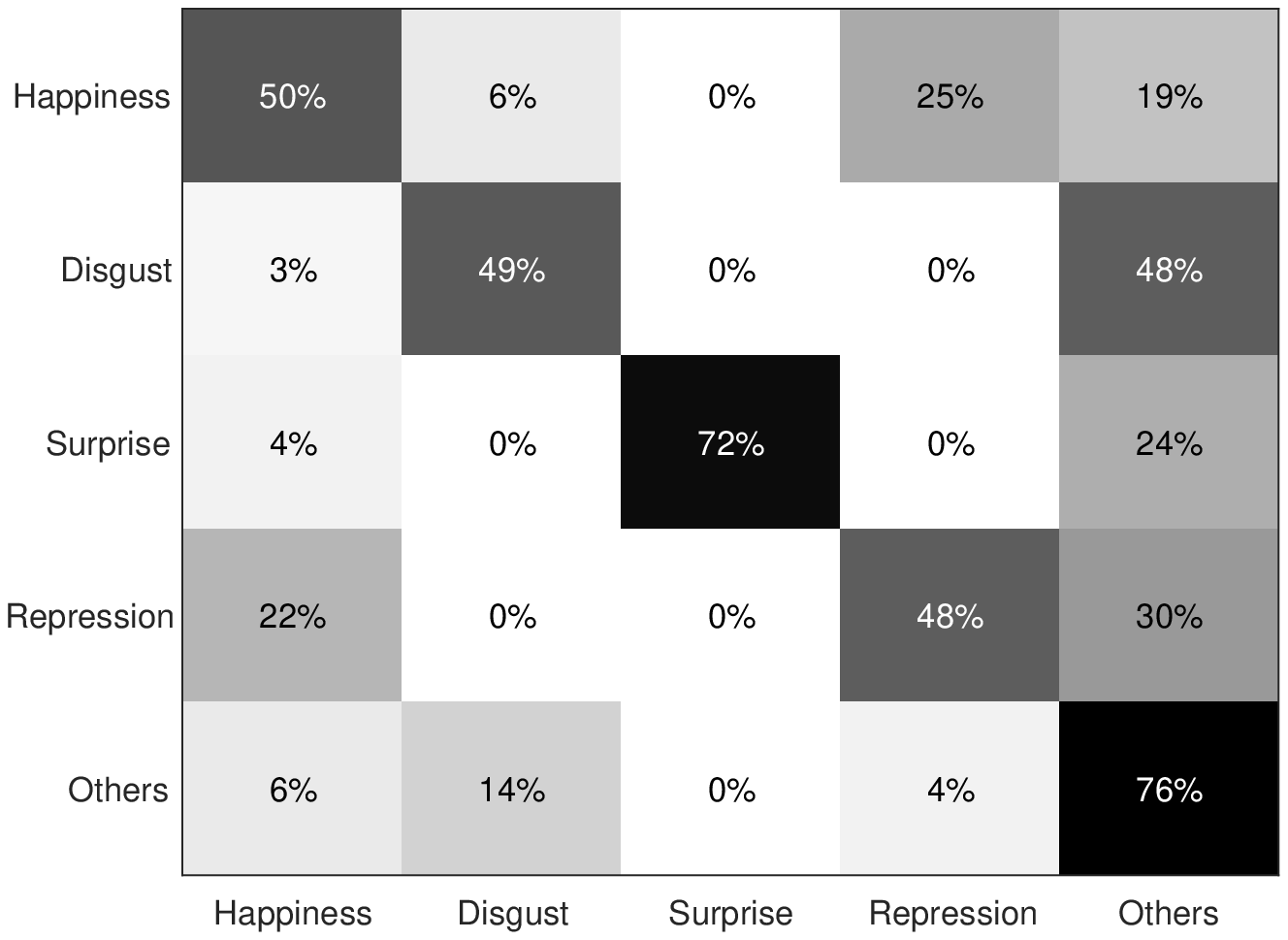}
}
\subfigure [LBP-SIP]{
\label{CSIP}
\includegraphics[height=5cm,width=7.8cm]{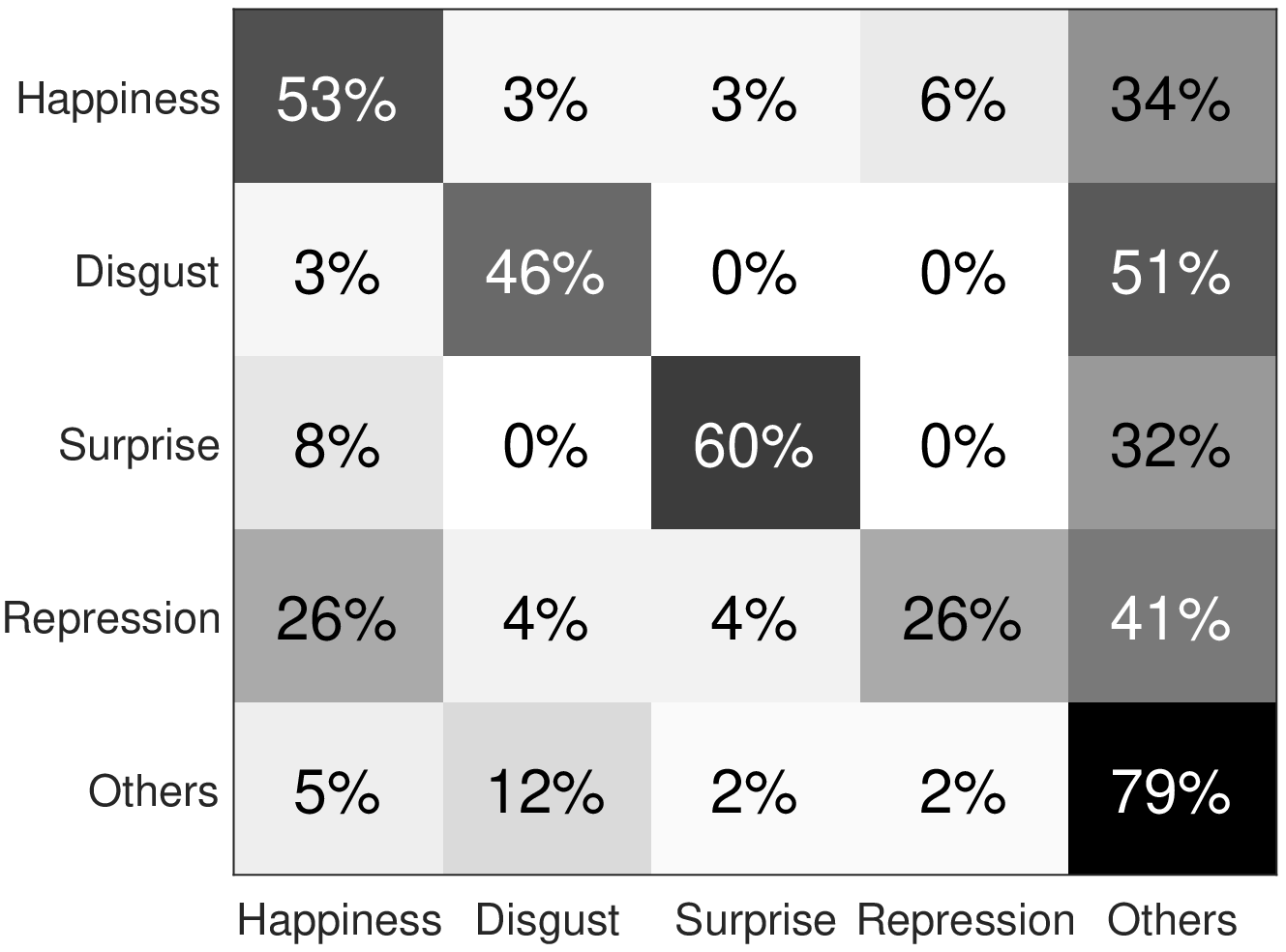}
}
\subfigure [SIP-SDG]{
\label{CSIP-SDG}
\includegraphics[height=5cm,width=7.8cm]{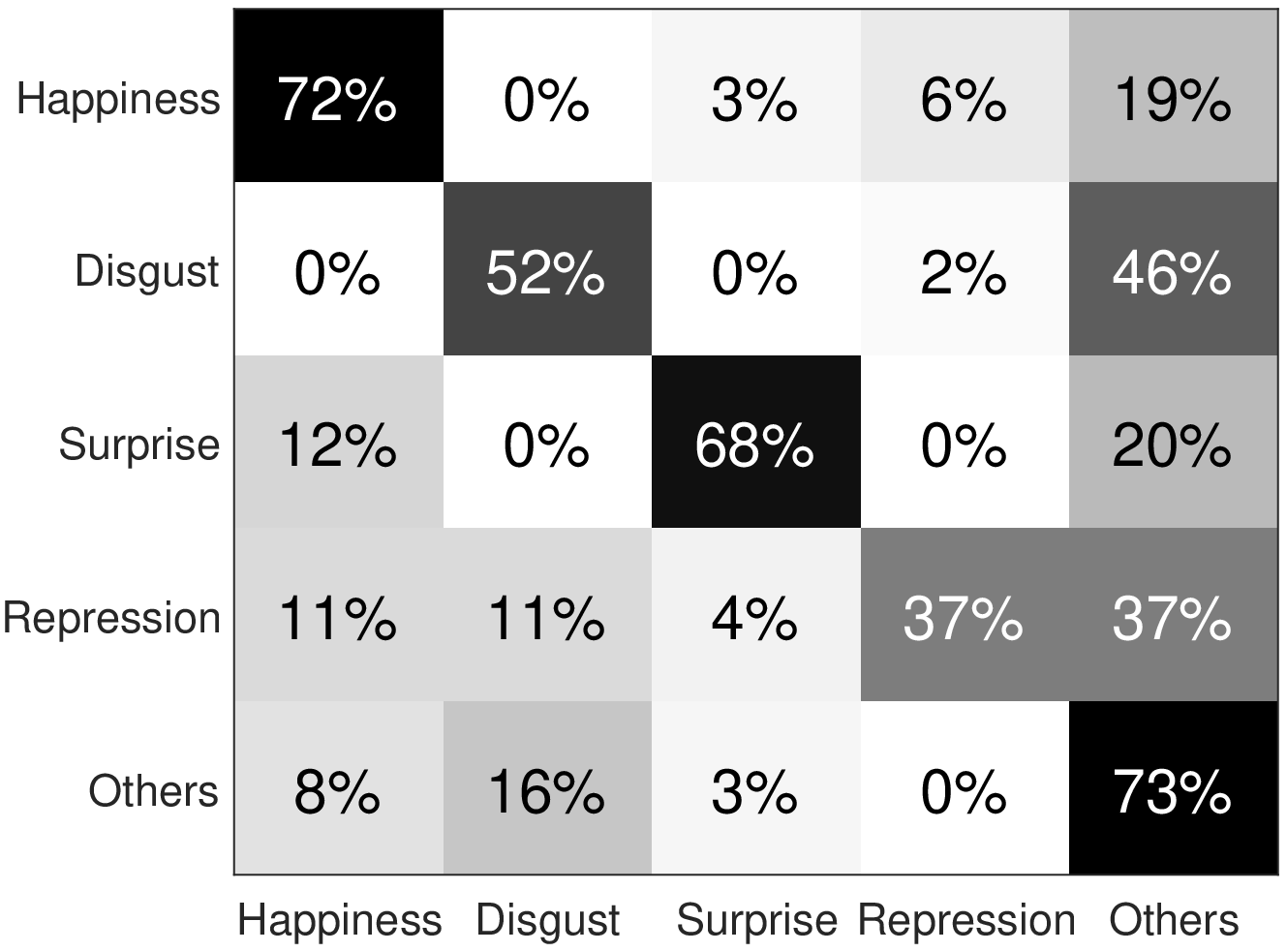}
}C
\subfigure [LBP-TOP.]{
\label{CTOP}
\includegraphics[height=5cm,width=7.8cm]{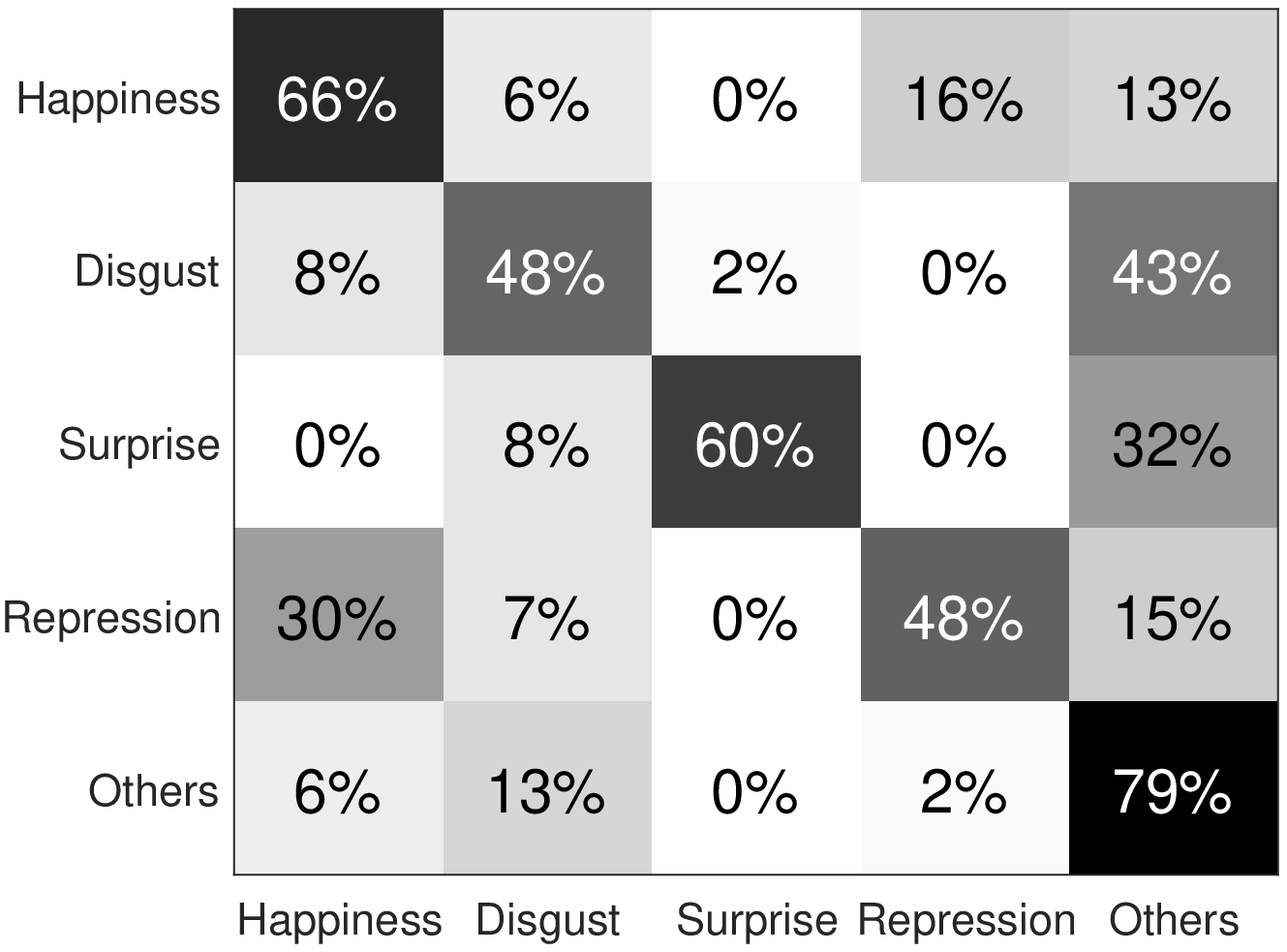}
}
\subfigure [LBP-SDG.]{
\label{CSDG}
\includegraphics[height=5cm,width=7.8cm]{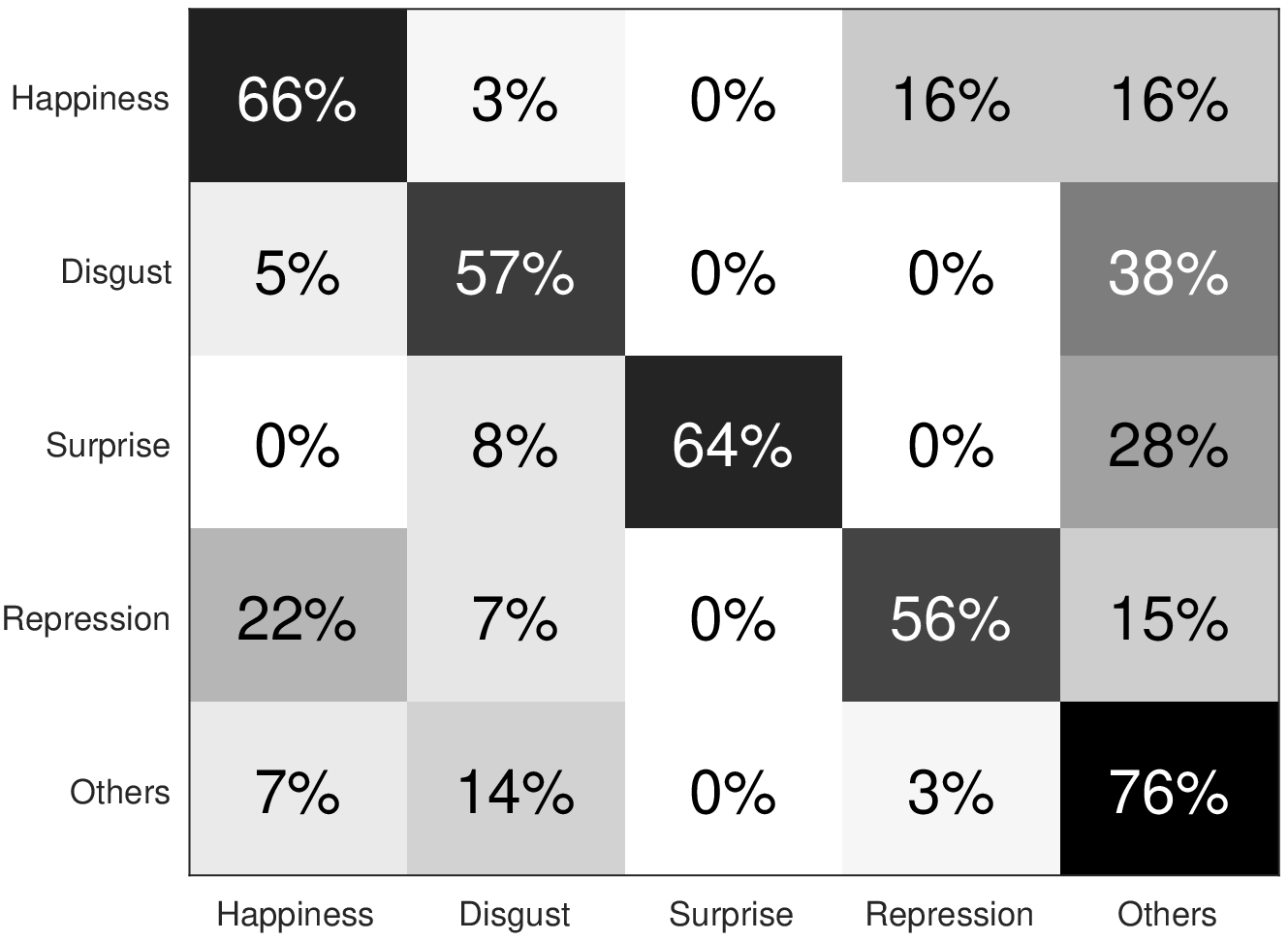}
}
\end{center}
\caption{The confusion matrices of different features on the CASME \uppercase\expandafter{\romannumeral2} database. }
\label{CConF}
\end{figure*}

\end{document}